%% file: main.tex
\documentclass[10pt,twocolumn,twoside]{IEEEtran}           % TWOCOLS

\usepackage{amsfonts}
\usepackage{amsbsy}
\usepackage{amsmath}
\usepackage{amsthm}
\usepackage{cite}
\usepackage{bbm}
\usepackage{breqn}
\usepackage{tikz}
\usepackage{subcaption}
\usepackage{bm}
\usepackage{multirow}
\usepackage{soul}
\usepackage{subcaption}
\usepackage[colorinlistoftodos]{todonotes}
\usepackage{flushend}

\newcommand{\imag}{\mathbbm{i}}

\newcommand{\mD}{\mathcal D}
\newcommand{\mH}{\mathcal H}

\newcommand{\X}{{\bf X}}
\newcommand{\x}{{\bf x}}

\newcommand{\y}{{\bf y}}
\newcommand{\w}{{\bf w}}
\newcommand{\M}{{\bf M}}
\newcommand{\U}{{\bf 1}}
\newcommand{\I}{{\bf I}}

\newcommand{\K}{{\bf K}}

\newcommand{\balpha}{\boldsymbol{\alpha}}
\newcommand{\bphi}{\boldsymbol{\phi}}
\newcommand{\bPhi}{\boldsymbol{\Phi}}
\newcommand{\bmu}{\boldsymbol{\mu}}

\newcommand{\Real}{\mathbb{R}}
\newcommand{\Comp}{\mathbb C}

\renewcommand{\baselinestretch}{1}

\newtheorem{definition}{\it Definition}
\newtheorem{theorem}{\it Theorem}

\usepackage{color}

\usepackage{hyperref}
\hypersetup{
    unicode=false,          % non-Latin characters in Acrobat’s bookmarks
    pdftoolbar=true,        % show Acrobat’s toolbar?
    pdfmenubar=true,        % show Acrobat’s menu?
    pdffitwindow=false,     % window fit to page when opened
    pdfstartview={FitH},    % fits the width of the page to the window
    pdftitle={DR},    % title
    pdfauthor={ALL},     % author
    pdfsubject={Nonlinear Distribution Regression in Remote Sensing},   % subject of the document
    pdfcreator={GCV}   % creator of the document
    pdfproducer={GCV}, % producer of the document
    pdfkeywords={}, % list of keywords
    pdfnewwindow=true,      % links in new window
    colorlinks=true,       % false: boxed links; true: colored links
    linkcolor={blue},          % color of internal links (change box color with linkbordercolor)
    citecolor=blue,        % color of links to bibliography
    filecolor=blue,      % color of file links
    urlcolor=blue           % color of external links
}

\title{Nonlinear Distribution Regression for\\Remote Sensing Applications}

\author{Jose E. Adsuara, Adri\'an P\'erez-Suay, Jordi Mu{\~n}oz-Mar\'i, Anna Mateo-Sanchis, \\
Maria Piles, Gustau Camps-Valls
\thanks{Image Processing Laboratory (IPL) \newline
Universitat de Val\`encia, Catedr\'atico A. Escardino - 46980 Paterna, Val\`encia (Spain). E-mail: gustau.camps@uv.es}
\thanks{Research funded by the European Research Council (ERC) under the ERC-CoG-2014 SEDAL project (grant agreement 647423) and the Spainish Ministry of Economy, Industry and Competitiveness IMINECO) under the `Network of Excellence' program (grant code TEC2016-81900-REDT), and MINECO and FEDER co-funding through project TIN2015-64210-R and RTI2018-096765-A-100.}
}

\begin{document}
\maketitle

\begin{abstract}
In many remote sensing applications one wants to estimate variables or parameters of interest from observations. When the target variable is available at a resolution that matches the remote sensing observations, standard algorithms such as neural networks, random forests or Gaussian processes are readily available to relate the two. 
However, we often encounter situations where the target variable is only available at the group level, i.e. collectively associated to a number of remotely sensed observations. This problem setting is known in statistics and machine learning as {\em multiple instance learning} or {\em distribution regression}. This paper introduces a nonlinear (kernel-based) method for distribution regression that solves the previous problems without making any assumption on the statistics of the grouped data.  
The presented formulation considers distribution embeddings in reproducing kernel Hilbert spaces, and performs standard least squares regression with the empirical means therein. A flexible version to deal with multisource data of different dimensionality and sample sizes is also presented and evaluated. It allows working with the native spatial resolution of each sensor, avoiding the need of match-up procedures. Noting the large computational cost of the approach, we introduce an efficient version via random Fourier features to cope with millions of points and groups.  
Real experiments involve SMAP Vegetation Optical Depth data for the estimation of crop production in the US Corn Belt, and MODIS and MISR reflectances for the estimation of Aerosol Optical Depth. An exhaustive empirical evaluation of the method is done against naive (linear and nonlinear) approaches based on input-space means, as well as previously presented methods for multiple-instance learning. 
We provide source code of our methods in \href{http://isp.uv.es/code/dr.html}{http://isp.uv.es/code/dr.html}.
\end{abstract}


%%%%%%%%%%%%%%%%%%%%%%%%%%%%%%%%%%%%%%%%%%%%%%%%%%%%%%%%%%%%%%%%%%%%%%%%%%%%%%%%%%%
\section{Introduction} \label{sec:1}
%%%%%%%%%%%%%%%%%%%%%%%%%%%%%%%%%%%%%%%%%%%%%%%%%%%%%%%%%%%%%%%%%%%%%%%%%%%%%%%%%%%

\IEEEPARstart{E}{stimating} variables and bio-geophysical parameters of interest from observations is a central problem in Earth observation~\cite{Liang08,rodgers00,CampsValls11mc}. From a statistical standpoint, the problem reduces to regression and function approximation, for which either physical, statistical or hybrid inversion techniques can be used~\cite{verrelst12b,CampsValls11mc}. In recent years, the amount of available data has allowed tackling complicated remote sensing regression problems learning. When input-output pairs are given, a plethora of algorithms can be readily used, such as neural networks~\cite{Ratle20102271}, random forests~\cite{Tramontana2015360} or kernel methods in general~\cite{Rojo17dspkm} and Gaussian processes in particular~\cite{CampsValls16grsm,warpedGP}, just to name a few. However, we often encounter problems where the target variable or parameter of interest is only available at the group level, not at the sample level. For example, very often one aims to estimate a bio-geophysical parameter, climate variable or ecological indicator from a set of observations, not just a single one, because of the spatial resolution or the acquisition level (e.g. ~\cite{verrelst12b}). Other common problems consider predicting a variable reported on field-based inventories or surveys at the county, region or state level from a set of observations (e.g. crop yield ~\cite{LopezLozano15} or forest biomass \cite{Galidaki2017}). The inverse situation, where an observation covers multiple samples, is also challenging. This is typically found in satellite retrieval validation, where coarse-scale estimates are compared to short-term intensive field campaign measurements, or to long-term -dense and sparse- station networks of point-scale ground based observations~\cite{Kucera2013,Crow2012,Tang2014}. 

Several strategies exist to address this input-output mismatch problem by either 1) {\em output expansion}, that is replicating the group label for all observations in the group, or alternatively by 2) {\em input summary} of all group feature vectors. The first approach makes the problem even more ill-posed, as all inputs in a group are associated with the same target variable. We should note here that, while this is actually a valid approach for classification, in regression problems labels have a semantic ordered meaning so `quantizing' the output space typically fails. When it comes to the second approach, it is customary to summarize all data in each group with its empirical mean or with a set of centroids (one per group) computed by clustering the group datasets. A more detailed description of these strategies are given below in Section~\ref{sec:2}.

While the previous approaches are quite convenient and intuitive, they introduce strong assumptions about the data distributions. 
First, selecting a good summarizing criterion for all points in a group (e.g. all pixels within a county) is far from trivial. The empirical mean assumes that only the first moment captures all variability inside a group and it is enough to distinguish between group target variables (e.g. crop yield), while clustering assumes a particular data distribution (e.g. via using the Euclidean distance) is valid for all groups. Obviously such assumptions do not necessarily hold in most real case studies. Following with the example of crop yield prediction, adopting the empirical mean strategy could lead to the same (or deemed similar) input average feature vector (spectral signature) for counties having completely different types of crops and areas planted. Secondly, by summarizing each set with an average input feature vector one implicitly assumes that all bags or grouped-pixels have the same relevance, despite the number and variability of observations in each one. It may happen, for example, that a smaller county (hence lower number of input spectra available) yields a similar average statistic as a larger county. This leads to systematically biased model estimations, to eventually skewed conclusions about accuracy and fit, and to potentially wrong uncertainty propagation analysis.

The distinct goal of {\em distribution regression} (DR) is that of exploiting all input data available without explicitly summarizing the groups or replicating target variables. Hence, one is eventually interested in regressing the variable against a distribution of input feature vectors, not just one summarizing vector.
Such a setup is ideally suited for the retrieval and validation of EO (Earth Observation) parameters characterized by high spatial variability that are measured in-situ at the point-scale. This is for instance the case of aerosols~\cite{Wang2008}, precipitation \cite{Kucera2013}, greenhouse gases~\cite{Charkovska2018}, soil moisture \cite{Crow2012}, land surface temperature~\cite{Moser2015}, ocean salinity~\cite{Tang2014}, among others. 
DR can also prove very useful for applications where remotely sensed observations are used to predict variables reported in surveys or inventories at a regional scale, especially in regions that are particularly heterogeneous in size and/or composition. This includes for instance prediction of crop yield, carbon stocks, vector-borne diseases, or insect plagues (e.g. desert locust). Other applications of DR may include detection of anomalies and changes of interesting phenomena in datasets, as not single observations but distributions of observations are exploited. In general, distribution regression is an ideally suited framework to tackle problems that need predicting a scalar value from a distribution.

In recent years, the problem of mapping distributions to point estimates has received a broad interest in statistics and machine learning, known under different names such as {\em area-to-point kriging} (ATPK)~\cite{Goovaerts2010CombiningAA}, {\em multiple instance learning} (MIL) for classification~\cite{Bolton11,Manandhar15,Jiao15,Yuksel15,Liu18} and regression~\cite{Wagstaff2007,Wang2008,Wang2012a}, yet the field has been recently formalized under the field of {\em distribution regression} (DR)~\cite{pmlr-v31-poczos13a,pmlr-v28-oliva13,pmlr-v33-oliva14a,Szabo2014,Law2017}. We place our proposal in this later field. 
This paper follows the principles in~\cite{Szabo2014} and introduces a nonlinear kernel-based DR method for remote sensing applications. The formulation considers distribution embeddings in reproducing kernel Hilbert spaces, and performs regression with the empirical kernel means therein. By virtue of the kernel trick, the explicit mean map embedding is not necessary to solve the problem or apply the final model to new input test data sets~\cite{Scholkopf02,Rojo17dspkm}. 

This paper presents a distribution regression framework specifically designed and adapted to the field of remote sensing. It introduces three main novelties. First, a version capable of dealing with multi-source data of different dimensionality and sample sizes is presented and evaluated. It allows working with the native spatial resolution of each sensor, something not allowed by any of the methods mentioned in the previous paragraph, avoiding the need of match-up procedures in which the information of the sensor with higher resolution is lost. Second, noting the large computational cost of the approach and the increasing amount of remote sensing data at high spatio-temporal resolutions available, we also introduce an efficient version via random Fourier features (RFF) that scales well to millions of points. Third, an extensive evaluation of the algorithm is provided to illustrate the applicability of the presented methodology 
using Soil Moisture Active Passive (SMAP) Vegetation Optical Depth for the prediction of crop yield over the US Corn Belt, and using Multi-angle Imaging Spectro-Radiometer (MISR) and Moderate Resolution Imaging Spectro-Radiometer (MODIS) reflectances from the TERRA satellite for the estimation of Aerosol Optical Depth. The method performs better than common approaches based on input-space means and clustering as well as previously presented methods for multiple-instance learning applied to similar datasets.

The rest of the paper is organized as follows. Section~\ref{sec:2} introduces the basic elements for nonlinear (kernel-based) distribution regression: kernel function and the mean map embedding, as well as related methods such as the maximum mean discrepancy to motivate the DR proposal. Then, in Section~\ref{sec:3}, we introduce our proposed kernel version for distribution regression, a multi-source version to deal with features of different dimensionality and sample sizes, and a fast version for computational efficiency based on random Fourier features. Experiments are detailed in Section~\ref{sec:4}. We conclude the paper in Section~\ref{sec:5} with some remarks and outline of the future work and opportunities.

%%%%%%%%%%%%%%%%%%%%%%%%%%%%%%%%%%%%%%%%%%%%%%%%%%%%%%%%%%%%%%%%%%%%%%%%%%%%%%%%%%%
\section{Kernel distribution regression} \label{sec:2}
%%%%%%%%%%%%%%%%%%%%%%%%%%%%%%%%%%%%%%%%%%%%%%%%%%%%%%%%%%%%%%%%%%%%%%%%%%%%%%%%%%%

% ---------------------------------------------------------------------------------
\subsection{Notation}
% ---------------------------------------------------------------------------------

Let us start by fixing the notation adopted in this paper. 
In distribution regression problems, we are given some sets of observations each of them with a corresponding output target variable to be estimated. Notationally, the training dataset $\mD$ is formed by a collection of $B$ bags (or sets) $\mD=\{(\X_b\in\Real^{n_b\times d},y_b\in\Real)|b=1,\ldots,B\}$. A training set from a particular group/bag $b$ is formed by $n_b$ examples, and is here denoted as $\X_b = [\x_1,\ldots,\x_{n_b}]^\top\in\Real^{n_b\times d}$, where $\x_i\in\Real^{d\times 1}$. For later convenience, let us denote all the available data collectively grouped in matrix $\X\in\Real^{n \times d}$, where $n=\sum_{b=1}^B n_b$, and $\y=[y_1,\ldots,y_B]^\top\in\Real^{B\times 1}$. This setting hampers the direct application of regression algorithms because not just a single input point $\x_b$ but a set of points $\X_b$ is available for model development, while latter for prediction we may have test points or sets from each bag denoted with a star superscript, $\x_b^*\in\Real^{d\times 1}$ or $\X_b^*\in\Real^{m_b\times d}$.

The problem in DR reduces to finding a function $f$ that learns the mapping from $\x$ to $y$, that is
$$y=f(\x)+e_i,~~~~e_i\sim{\mathcal N}(0,\sigma_n^2).$$
Such a model, however, poses a challenging situation for standard regression since $\mD$ contains many-to-one data. As discussed before, two main approaches are typically followed: 1) {\em output expansion}, that is replicating the label $y_b$ for all points in bag $b$; or 2) {\em input summary} most notably with the empirical average $\bar\x_b = \frac{1}{n_b}\sum_i \x_i$, or a set of centroids ${\bf c}_b$, $b=1,\ldots,B$. The distinct goal of {\em distribution regression} is to exploit the rich structure in $\mD$ by performing regression with the group distributions directly. Statistically, this boils down to exploit all higher order statistical relationships between the groups, not just the first or second order moments. In this paper we present a method that embeds the bag distribution in a Hilbert space and performs linear regression therein. Some tools from the theory of reproducing kernels and functional analysis are needed, which are reviewed in what follows.

% ---------------------------------------------------------------------------------
\subsection{Kernels functions and the mean map embedding}\label{subsec:MMD} 
% ---------------------------------------------------------------------------------

In this work we will rely on the theory of kernel methods to tackle the distribution regression problem. Let us summarize briefly the main needed concepts: kernel function, feature map, representer theorem, mean map embedding and the kernel ridge regression as the method used for estimation. For more comprehensive explanations the reader is addressed to the books~\cite{Scholkopf02,ShaweTaylor04,Rojo17dspkm,CampsValls09}. A detailed treatment of kernel mean embedding of distributions can be found in~\cite{Muandet2016}.

% ---------------------------------------------------------------------------------
\subsubsection{Kernel functions and feature maps}
% ---------------------------------------------------------------------------------

Kernel methods rely on the notion of similarity between samples in a higher (possibly infinite dimensional) Hilbert space. Be a set of empirical data $\x_1,\ldots,\x_n\in{\mathcal X}$, where the points are defined in a $d$-dimensional input space, $\x=[x^1, \ldots,x^d]^\top\in\Real^d$. Kernel methods assume the existence of a (dot product) Hilbert space ${\mathcal H}$, where samples are mapped into by means of a feature map $\bphi:{\mathcal X}\rightarrow{\mathcal H}, \x\mapsto \bphi(\x)$. The mapping function can be defined explicitly (if some prior knowledge about the problem is available) or implicitly which is often the case in kernel methods. The similarity between the elements in ${\mathcal H}$ can now be measured using its associated dot product $\langle\cdot,\cdot\rangle_{{\mathcal H}}$. Here, we define a function that computes that similarity kernel, $k:{\mathcal X}\times{\mathcal X}\rightarrow\mathbb{R}$, such that $(\x,\x')$ $\mapsto$ $k(\x,\x')$. This function, often simply called kernel, is required to satisfy the Mercer's Theorem~\cite{Scholkopf02}: 
\begin{equation}\label{eq:kernel}
  k(\x,\x') = \langle \bphi(\x), \bphi (\x') \rangle_{{\mathcal H}}.
\end{equation}
The mapping $ \bphi$ is its \emph{feature map}, the space ${\mathcal H}$ is the reproducing Hilbert \emph{feature space}, and $k$ is the reproducing kernel function since it {\em reproduces} dot products in ${\mathcal H}$ without even mapping the data explicitly therein. The point in kernel methods theory is that one relies on the implicit definition of the kernel function, and hence no feature mapping is explicitly defined. In all our experiments we used the radial basis function (RBF) kernel function, which actually accounts for all higher order monomials. Let us assume the kernel $k(x,x')=(\phi(x),\phi(x'))=\exp(-\gamma(x-x')^2)$, where for simplicity we define $\gamma>0$. Then, the explicit feature map $\phi(x)$ is infinite dimensional, and can be expressed as 
\begin{equation*}\label{eq:expKernel}
\phi(x)=\exp(-\gamma x^2) \bigg[1,\sqrt{\frac{2\gamma}{1!}}x,\sqrt{\frac{(2\gamma)^2}{2!}}x^2,\sqrt{\frac{(2\gamma)^3}{3!}}x^3,...\bigg]
\end{equation*}
and hence all higher-order relations between $x$ and $x'$ through the consideration of all monomials in the dot product reproduced by the kernel. Figure~\ref{mmd_performance}(d) shows a concrete illustrative example of detection of differences between distributions by using an explicit map  $\phi(x)=x^2$. In kernel methods, however, the main advantage is that the mapping does not need to be explicitly designed.

\begin{definition}{Reproducing kernel Hilbert spaces (rkHs)~\cite{Aronszajn50}.}
A Hilbert space $\mathcal{H}$ is said to be a rkHs if: (1) The elements of $\mathcal{H}$ are complex or real valued functions $f(\cdot)$ defined on any set of elements $\x$; And (2) for every element $\x$, $f(\cdot)$ is bounded.
\end{definition}
The name of these spaces come from the so-called {\em reproducing property}. Indeed, in an rkHs ${\mathcal{H}}$, there exists a function $k(\cdot,\cdot)$ such that 
\begin{equation}\label{rkHs_property}
f(\x)=\langle f(\cdot),k(\cdot,\x) \rangle,~~~f \in \mathcal{H}
\end{equation}
by virtue of the Riesz Representation Theorem~\cite{RieNag55}. A large class of algorithms has originated from regularization schemes in rkHs. The representer theorem gives us the general form of the solution to the common loss formed by a cost (loss, energy) term and a regularization term.

{\theorem{(Representer Theorem)~\cite{Kimeldorf1971}}\label{representer}
Let $\Omega:[0,\infty)\rightarrow\mathbb{R}$ be a strictly monotonic
increasing function; let 
$V:\mathbb{R}\times\mathbb{R}\rightarrow\mathbb{R}\cup\{\infty\}$
be an arbitrary loss function; and let ${\mathcal H}$ be a rkHs with reproducing kernel $k$. Then:
\begin{equation}\label{funct_representer}
f^\ast = \min_{f\in{\mathcal H}}\bigg\{V\left((f(\x_1), y_1),\ldots,(f(\x_n),y_n)\right)+\Omega(\|f\|^2_{{\mathcal H}})\bigg\}
\end{equation}
admits a space of functions (representation) $f$ defined as
\begin{equation}\label{eq:representation}
f(\x) = \sum_{i=1}^n\alpha_i k(\x,\x _i),~~~\alpha_i\in\mathbb{R},~~\balpha \in\Real^{n\times 1}
\end{equation}
This is, the function that minimizes the (regularized) optimization functional in Eq.~\eqref{funct_representer} is a linear combination of dot products between data mapped into ${\mathcal H}$.}

% ---------------------------------------------------------------------------------
\subsubsection{Mean map embeddings}
% ---------------------------------------------------------------------------------

{We frame the problem in the theory of mean map embeddings of distributions~\cite{Rojo17dspkm,Harchaoui13,Muandet2016}. Let ${\mathcal B}_{\mathcal X}$ be the set of all probability distributions, then the kernel mean map $\bmu$ is defined as
$$\bmu:{\mathcal B}_{\mathcal X}\to {\mathcal H},~~~{\mathbb P}\to\int_{\mathcal X} k(\cdot,\x)\text{d}{\mathbb P}(\x)\in{\mathcal H}.$$
{Assuming that $k(\cdot,\x)$ is bounded for any $\x \in {\mathcal X}$, we can show that for any ${\mathbb P}$, letting $\bmu_{\mathbb P} = \bmu({\mathbb P})$, the ${\mathbb E}_P[f] = \langle \bmu_{\mathbb P},f\rangle_{\mathcal H}$, for all $f\in {\mathcal H}$}. It is important to note that, as for rkHs, here $\bmu$ represents the expectation function on ${\mathcal H}$. Following \cite{Muandet2016}, every probability measure has a unique embedding and the $\bmu$ fully determines the corresponding probability measure.}

{Now the question is how to estimate such mean embeddings from empirical samples. Following our notation before for one particular bag, $\X_b$, drawn i.i.d. from a particular ${\mathbb P}_b$, the empirical mean estimator of $\bmu_b$ is given by:
\begin{equation}
\widehat\bmu_b = \bmu_{{\mathbb P}_b} = \int k(\cdot,\x)\hat{\mathbb P}(\text{d}\x) \approx \dfrac{1}{n_b}\sum_{i=1}^{n_b}k(\cdot,\x_i).\label{eq:dotmeans}
\end{equation}
This is an empirical mean map estimator whose dot product can be computed via kernels:
\begin{equation}
\langle\widehat\bmu_{{\mathbb P}_b},\widehat\bmu_{{\mathbb P}_{b'}}\rangle_{\mathcal H} = \dfrac{1}{n_bn_{b'}} \sum_{i=1}^{n_b}\sum_{j=1}^{n_{b'}} k(\x_i^b,\x_j^{b'}).
\end{equation}
}

\begin{figure*}[t]
\begin{center}
\setlength{\tabcolsep}{1pt}
\begin{tabular}{cccc}
(a) & (b) & (c) & (d) \\
\includegraphics[height=3.5cm]{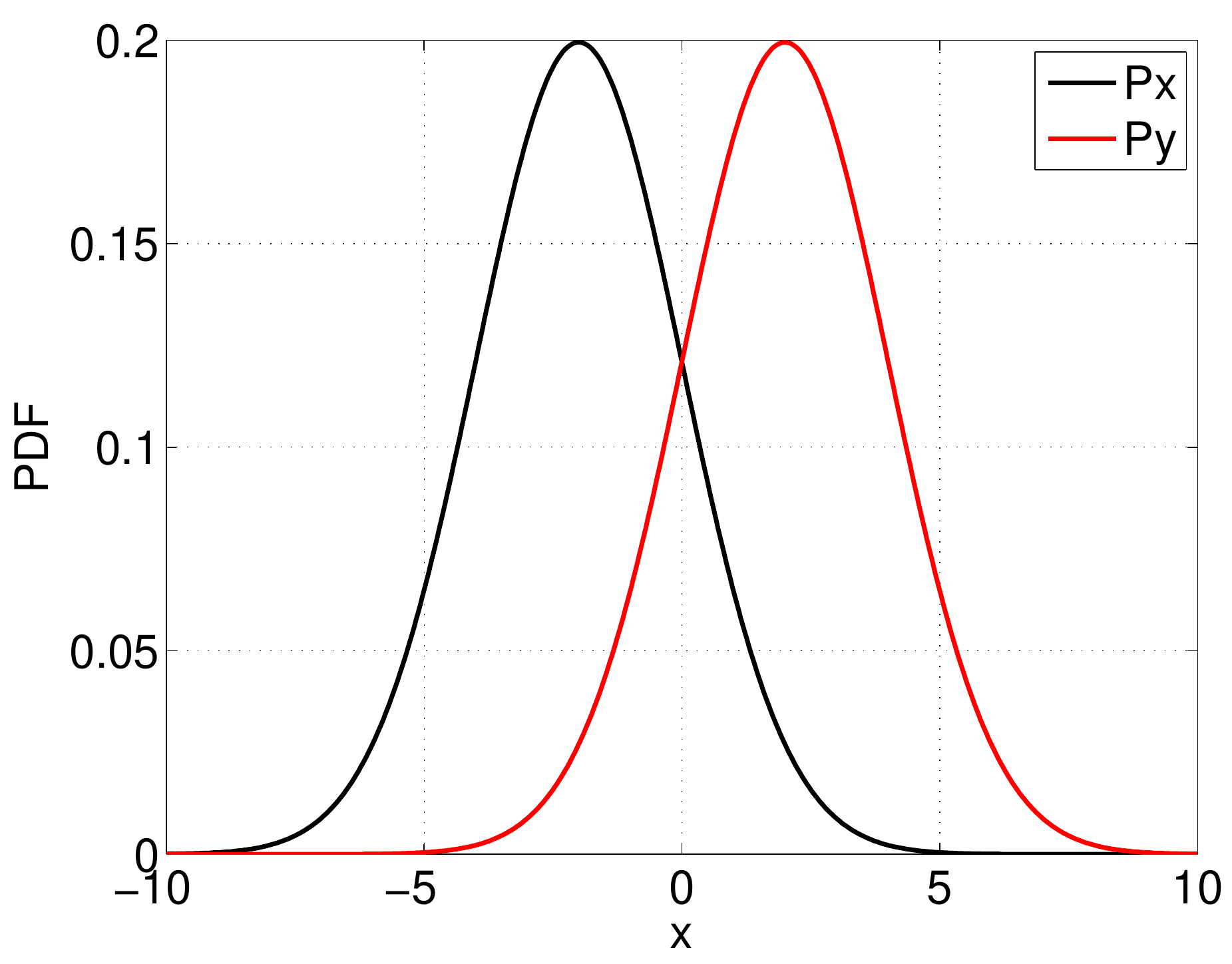} &
\includegraphics[height=3.5cm]{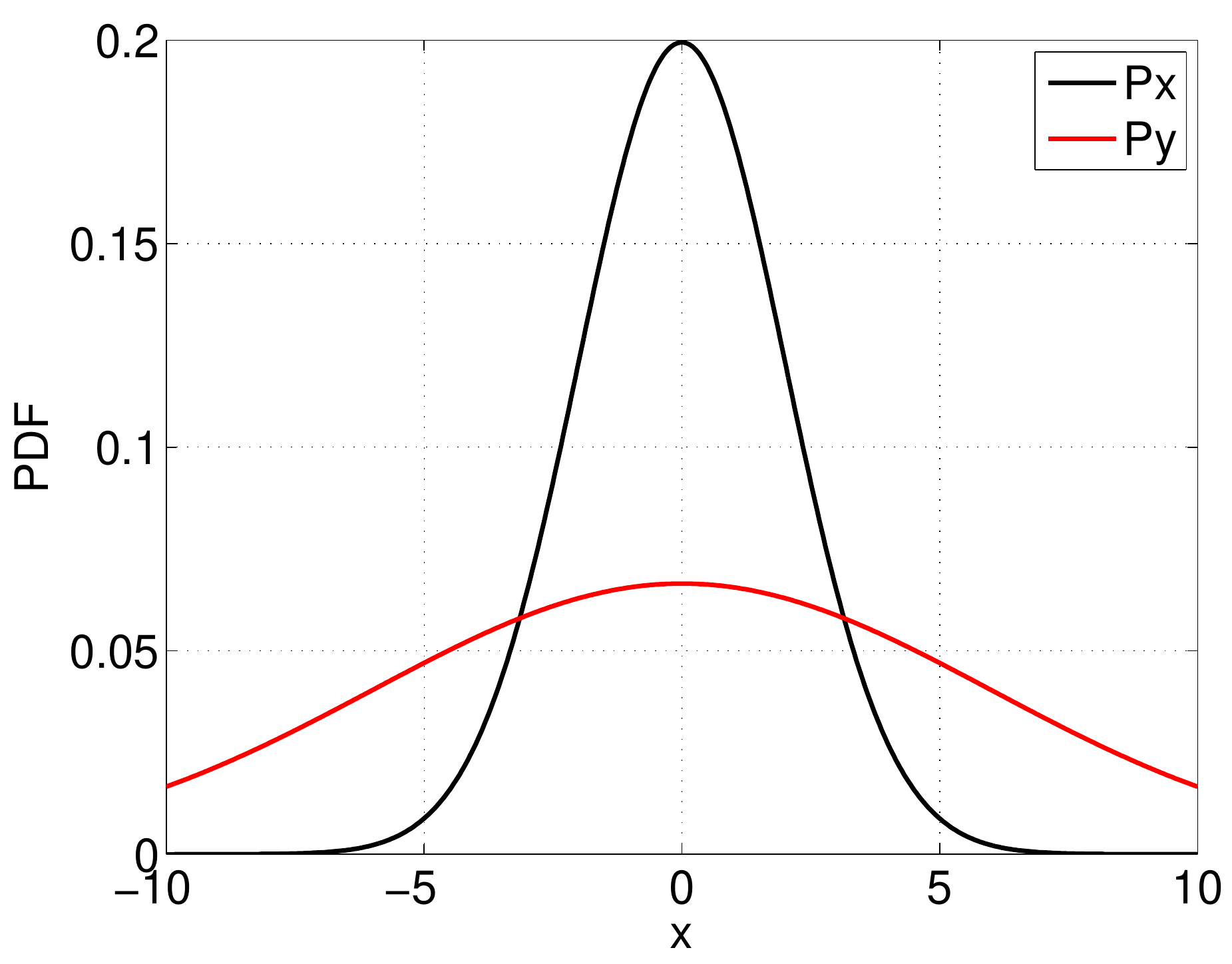} &
\includegraphics[height=3.5cm]{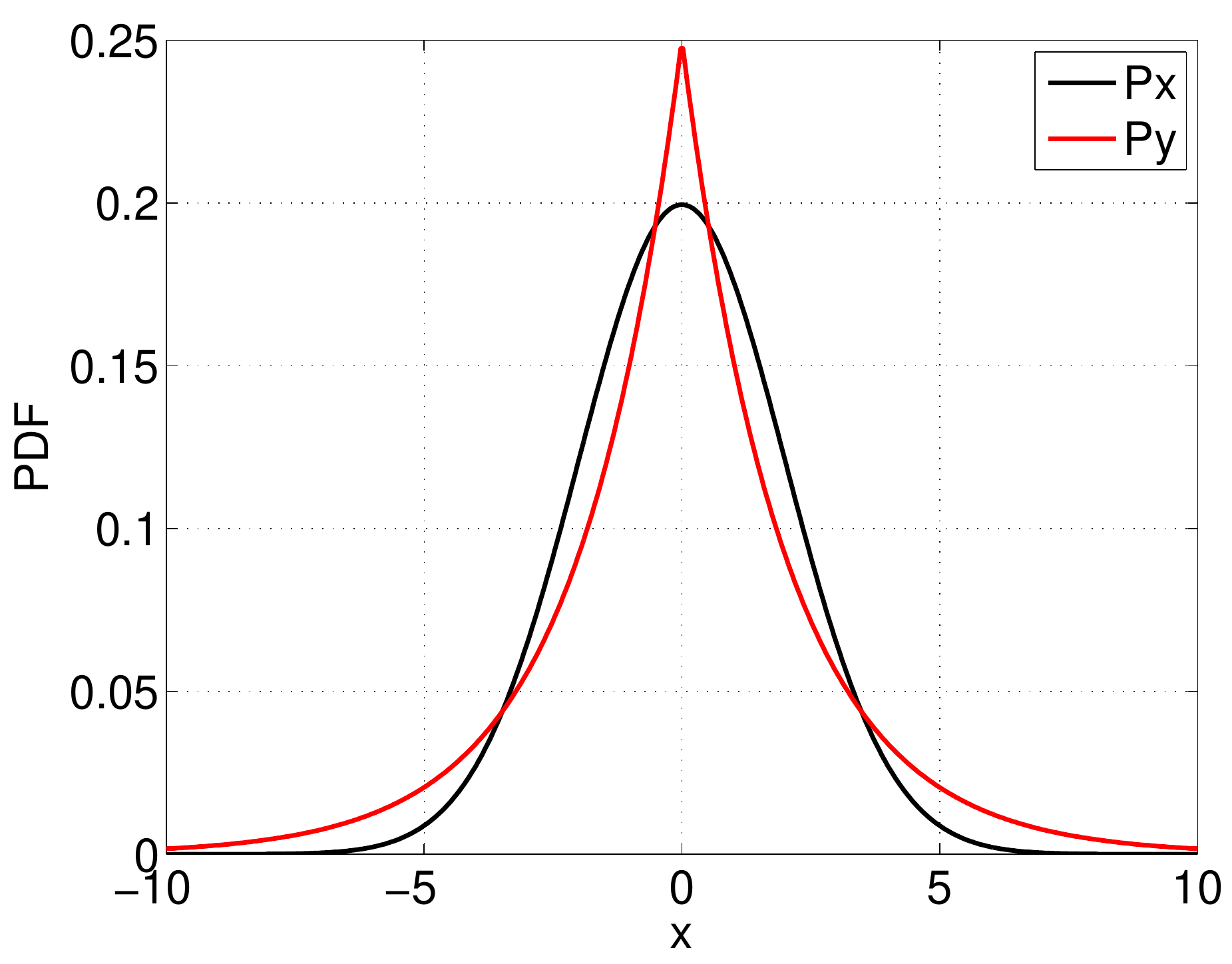} &
\includegraphics[height=3.5cm]{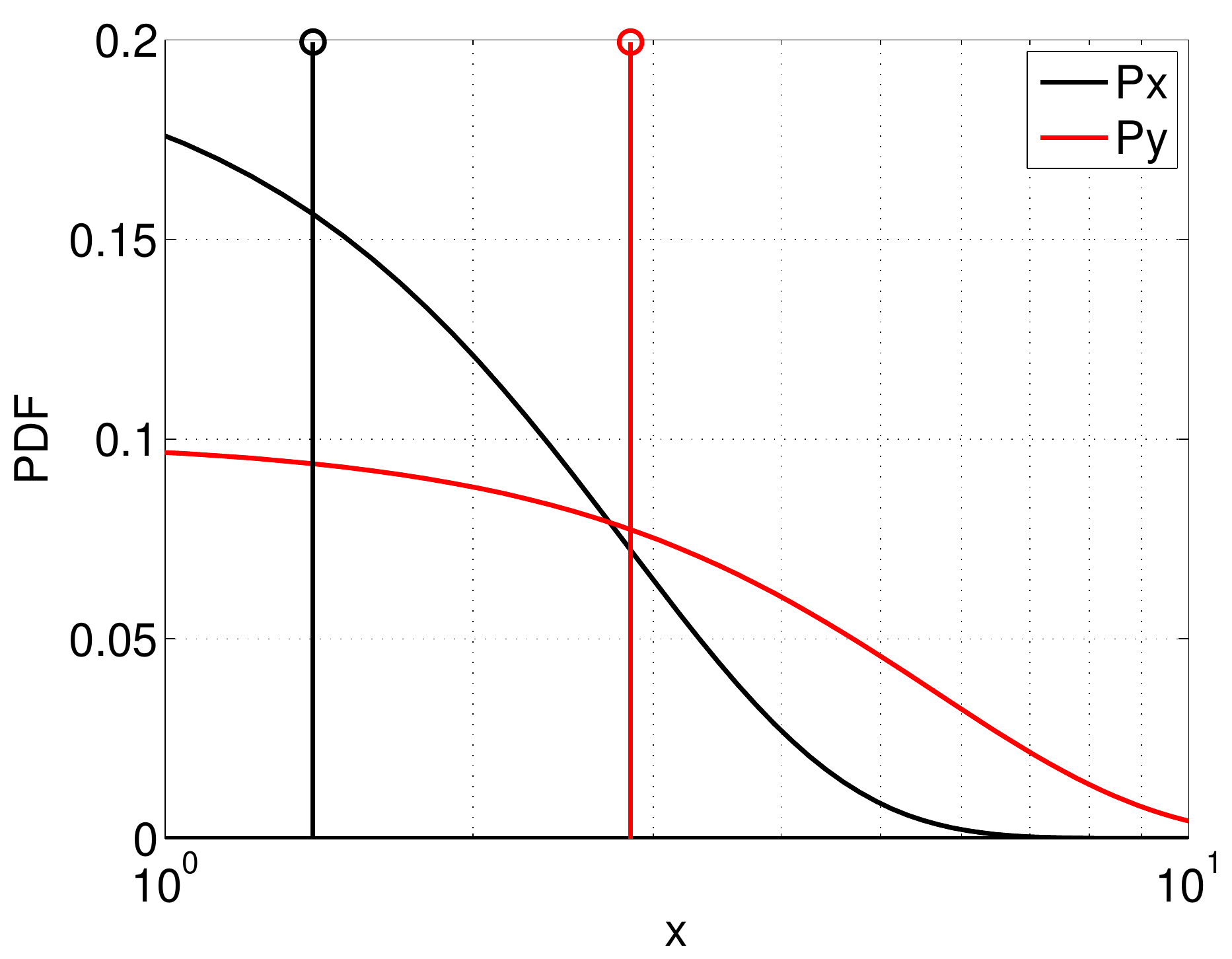} \\
\end{tabular}
\end{center}
\vspace{-0.25cm}
\caption{The two-sample problem reduces to detecting whether two distributions ${\mathbb P}_x$ and ${\mathbb P}_y$ are different or not. Summarizing group distributions for regression using a statistic is common practice, but it may fail because completely different groups can be indistinguishable and make regression strongly ill-posed. (a) Whenever we have two Gaussians with different means one can assess statistical differences by means of a standard $t$-test of different means (the distance between the empirical means is $d_{xy}:=\|\mu_x-\mu_y\|=3.91$) and summarizing the distributions with the empirical mean is good enough; (b) when we have two Gaussians with the same mean but different variance using only the first moment is useless ($d_{xy}=0.02$), yet one could resort to the second moment -the variance- which is what the $z$-test does; (c) however, we may find situations with the same first and second order yet different in nature (in this case, Gaussian and Laplace distributions with the same mean and variance), which hampers again discrimination and the definition of the summary statistic for regression; (d) the latter can be easily addressed by mapping the random variables to higher order features for discrimination (e.g. for the Gaussian case, second order features of the form $x^2$ suffice for discrimination, $d_{xy}=13.26$).
This simple example motivates the use of kernel mean embeddings for distribution regression and hypothesis testing that are able to estimate all higher order moments without even mapping the data explicitly but resorting to kernel functions only. \label{mmd_performance}}
\end{figure*}

Actually one can compute also distances among mean embeddings, which results in a particularly useful kernel algorithm for hypothesis testing and domain adaptation named Maximum Mean Discrepancy (MMD)~\cite{Muandet2016,Rojo17dspkm}, which has been previously used in remote sensing for feature selection~\cite{CampsValls10hsic}, classification and domain adaptation~\cite{Matasci15,Persello16}. In both hypothesis testing and distribution regression we are ultimately concerned about comparing distributions ${\mathbb P}_b$ and ${\mathbb P}_{b'}$. MMD reduces to estimating the distance between the two sample means in a reproducing kernel Hilbert space ${\mathcal H}$ where data are embedded
$$\text{MMD}({\mathbb P}_b,{\mathbb P}_{b'}) := \|\bmu_{{\mathbb P}_b} - \bmu_{{\mathbb P}_{b'}} \|^2_{\mathcal H},$$
which can be computed with kernels exploiting the dot product in Eq.~\eqref{eq:dotmeans}. Interestingly, MMD tends asymptotically to zero when the two distributions ${\mathbb P}_x$ and ${\mathbb P}_y$ are the same, which allows us to assess differences in distributions of possibly different nature from empirical samples. Figure~\ref{mmd_performance} illustrates the ability of MMD and the mean map embeddings to detect such differences hidden in higher order statistics, and motivates the use of kernel embeddings for distribution regression.

% ---------------------------------------------------------------------------------
\subsubsection{Kernel ridge regression} 
% ---------------------------------------------------------------------------------

Finally, let us now formally define the distribution regression task. For this we perform standard least squares regression using the mean embedded data in Hilbert spaces. As we will see, the solution leads to that of the kernel ridge regression (KRR) algorithm~\cite{ShaweTaylor04} working with mean map embeddings. In our setting we want to minimize a classical regularized functional composed of two terms: the least square errors of the approximation (of the mean embedding) and a regularizer over the class of functions to be learned in Hilbert space $f\in{\mathcal H}$:
$$f^* = \arg\min_{f\in{\mathcal H}} \bigg\{\dfrac{1}{n}\sum_{i=1}^n\|y_i-f(\bmu_i)\|^2 + \lambda\|f\|^2_{\mathcal H}\bigg\},$$
where $\lambda>0$ is the regularization term. The ridge regression objective function has an analytical solution for a test set given a set of training examples:
$$\hat f_{\bmu_t} = {\bf k}(\K+n\lambda{\bf I})^{-1}\y,$$
where $\bmu_t$ is the mean embedding of the test set $\X_t$, ${\bf k}=[k(\bmu_1,\bmu_t),\ldots,k(\bmu_n,\bmu_t)]^\top\in\Real^{n\times 1}$, $\K=[k(\bmu_i,\bmu_j)]\in\Real^{n\times n}$ and $\y=[y_1,\ldots,y_n]^\top$ collectively gathers all outputs.

%%%%%%%%%%%%%%%%%%%%%%%%%%%%%%%%%%%%%%%%%%%%%%%%%%%%%%%%%%%%%%%%%%%%%%%%%%%%%%%%%%%
\section{Proposed kernel distribution regression} \label{sec:3}
%%%%%%%%%%%%%%%%%%%%%%%%%%%%%%%%%%%%%%%%%%%%%%%%%%%%%%%%%%%%%%%%%%%%%%%%%%%%%%%%%%%

% ---------------------------------------------------------------------------------
\subsection{Formulation}\label{subsec:DR}
% ---------------------------------------------------------------------------------

Let us first define a feature map that takes input samples into a Hilbert space, $\bphi:\x\in\Real^{H\times1}\to\bphi(\x)\in\Real^{H\times 1}$. All mapped data can be collectively grouped in a data matrix in Hilbert space denoted as $\mH$: $\bPhi_b\in\Real^{n_b\times H}$. The goal in kernel distribution regression is to perform a linear regression in ${\mathcal H}$. In order to do this we can summarize the bag feature vectors with the mean map embedding of samples in bag $b$, which is denoted here as $\bmu_b = \frac{1}{n_b}\sum_{i=1}^{n_b}\bphi(\x_i^b) \in \mH\in\Real^{H\times 1}$. Now, the collection of all mean embeddings of data (in the same $\mH$) is expressed as $\M=[\bmu_1|\cdots|\bmu_B]^\top\in\Real^{B\times H}$. 

At this point we aim to perform regression on mean embeddings. We define a linear regression model 
$$\hat y_b=\bmu_b^\top\w,~~b=1,\ldots,B,$$ 
where model weights $\w\in\Real^{H\times 1}$ are shared across all bag models. For the sake of convenience, we can define the predictive model in matrix form as $\hat \y = \M \w.$ %where $\times 1 = (B\times H) \cdot (H\times 1).$
The least squares solution corresponds to the normal (Wiener-Hopf) equation, $\w=\M^\dag\y = (\M^\top\M + \lambda\I)^{-1}\M^\top\y$. Nevertheless, noting the high dimensionalty of matrix $\M$, and hence the covariance matrix $\M^\top\M$, the problem cannot be explicitly solved. In order to address this, we follow the standard procedure in kernel methods by which we define a kernel function reproducing a dot product in $\mH$, $k(\x_i,\x_j)=\bphi(\x_i)^\top\bphi(\x_j)$ and a representer theorem for model weights, that is $\w = \sum_{b=1}^B \alpha_b \bmu_b = \M^\top\balpha$, where $\balpha=[\alpha_1,\ldots,\alpha_B]\in\Real^{B\times 1}$. Now, the dual solution is given by $\balpha = (\widetilde\K + \lambda\I)^{-1}\y$ with $\widetilde\K = \M\M^\top\in\Real^{B\times B}$, which can be readily recognized as the kernel ridge regression working on a kernel with entries: 
\begin{eqnarray*}
\begin{array}{ll}
[\widetilde\K]_{b,b'} &= \bmu_b^\top\bmu_{b'} = \dfrac{1}{n_b n_{b'}} \sum_{i=1}^{n_b}\sum_{j=1}^{n_{b'}}\bphi(\x_i^b)^\top\bphi(\x_j^{b'})  \\[4mm]
&= \dfrac{1}{n_b n_{b'}} \sum_{i=1}^{n_b}\sum_{j=1}^{n_{b'}}k(\x_i^b,\x_j^{b'}) = \dfrac{1}{n_b n_{b'}}\U_{n_b}^\top \K_{bb'} \U_{n_{b'}},
\end{array}
\end{eqnarray*}
where the matrix $\K_{bb'}\in\Real^{n_b\times n_{b'}}$. 
Therefore, we have an analytic solution of the problem. Now the question is how one can generate predictions for particular samples. Let us define a test sample matrix from a particular bag denoted as $\X_b^* = [\x_1,\ldots,\x_{m_b}]^\top\in\Real^{m_b\times d}$, which has a mean embedding $\bmu_b^* = \frac{1}{m_b}\sum_{l=1}^{m_b}\bphi(\x_l^b) \in \mH\in\Real^{H\times 1}$. The predictions for the test data can be explicitly derived as follows: 
\begin{align}
&\hat y_b^* = {\bmu_b^*}^\top\w  = \frac{1}{m_b}\sum_{l=1}^{m_b}\bphi^\top(\x_l^b)~\M^\top\balpha \\ 
           & =  \frac{1}{m_b}\sum_{l=1}^{m_b}\bphi^\top(\x_l^b)~[\bmu_1|\cdots|\bmu_B]\balpha  \nonumber \\ 
           & = \frac{1}{m_b}\sum_{l=1}^{m_b}\bphi^\top(\x_l^b)~\bigg[\frac{1}{n_1}\sum_{i=1}^{n_1}\bphi(\x_i^1) \bigg| \cdots \bigg| \frac{1}{n_B}\sum_{i=1}^{n_B}\bphi(\x_i^B)\bigg]\balpha  \nonumber\\
           & = \bigg[\frac{1}{m_b n_1}\sum_{l=1}^{m_b}\sum_{i=1}^{n_1}k(\x_l^b,\x_i^1), \cdots , \frac{1}{m_bn_B}\sum_{l=1}^{m_b}\sum_{i=1}^{n_B}k(\x_l^b,\x_i^B)\bigg]\balpha  \nonumber\\
           & = \frac{1}{m_b}\sum_{b'=1}^B \frac{1}{n_{b'}} \alpha_{b'} \sum_{l=1}^{m_b}\sum_{i=1}^{n_b}k(\x_l^b,\x_i^{b'})  = \frac{1}{m_b n}\U_{m_b}^\top\K_{bb'}\U_{n_{b'}}\balpha, \nonumber
\end{align}
where $\K_{bb'}\in\Real^{m_b\times n_{b'}}$ which is computed easily given a valid (Mercer) kernel function $k$.

% ---------------------------------------------------------------------------------
\subsection{Multisource distribution regression}\label{sec:mdr}
% ---------------------------------------------------------------------------------

Assume that now we aim to combine/exploit multi-modal (multi-source) information defining each bag. The sets may have different numbers of both features and sizes, e.g. we aim to combine different spatial, spectral or temporal resolutions. We will illustrate this case in the experimental section by combining at a bag level two distinct sensors for aerosol parameter estimation. 

\begin{figure*}[t!]
\hspace{.17\textwidth} (a) \hspace{5cm} (b)  \hspace{5cm}(c)
\vspace{-.4cm}
\begin{center}
\setlength{\tabcolsep}{2pt}
\begin{tabular}{ccc}
\multicolumn{3}{c}{\includegraphics[width=\textwidth]{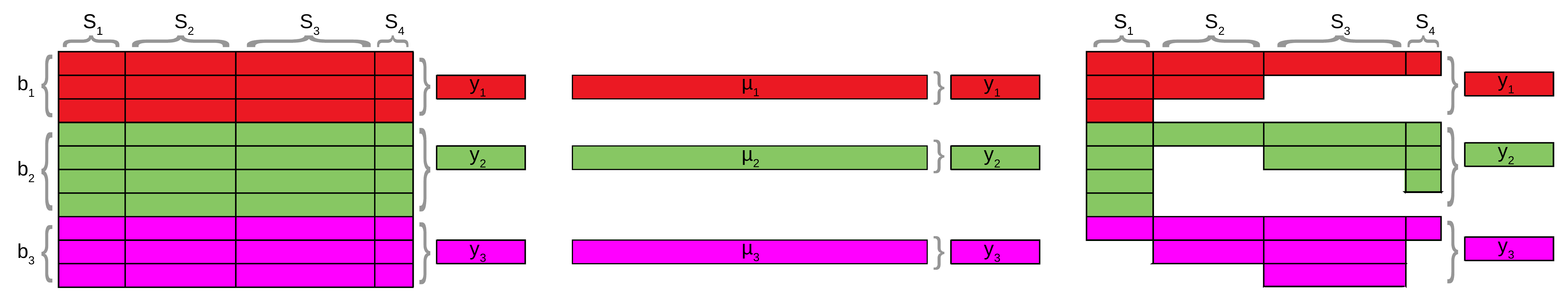}} \\
\end{tabular}
\end{center}
\vspace{-0.25cm}
\caption{Distribution regression approaches exemplified. The general DR problem setting is illustrated in (a), cf. \S\ref{subsec:DR}. Consider $B=3$ bags (different colors: red, green and purple) with different number of samples per bag ($n_1=3$, $n_2=4$, $n_3=3$) and three corresponding target labels, $y_b$, $b=1,2,3$. Columns are different sources of information, in this case there are four sensors acquiring data $S_i$, $i=1,2,3,4$. Note that the input feature space for each sensor $S_i$ does not necessarily have the same dimensionality (e.g. spectral or temporal resolution). Standard practice summarizes the distributions $\mathbb{P}_b$ with the mean vector $\mu_b$, $b=1,2,3$, where $\mu_b = [\mu_b^{S_1},\mu_b^{S_2},\mu_b^{S_3},\mu_b^{S_4}]$ (b), and then proceed with standard regression methods. This can be done in Hilbert spaces too but with the advantage of considering all moments of the distributions, not just the first one, and accounting for the relations among bag samples (cf. \S\ref{subsec:MMD}). In (c) we show the case of multi-source distribution regression (MDR) in which some features are missing for particular bags and samples, which is often encountered when different sensors are combined. This case is addressed in this paper too (cf. \S\ref{sec:mdr}).}
\label{fig:boxexamples}
\end{figure*}

Notationally, now we have access to different matrices $\X_f^b\in\Real^{n_b^f\times f}$, $f=1,\ldots,F$. We propose a multimodal kernel distribution method by embedding each dataset into a mean and exploiting the direct sum of Hilbert spaces in the mean embedding space. To do this, we define $F$ Hilbert spaces ${\mathcal H}_f$, $f=1,\ldots,F$, and the direct sum of all of them, ${\mathcal H} = \bigoplus_{f=1}^F {\mathcal H}_f$. Now the key for multimodal distribution regression is to perform a linear regression in ${\mathcal H}$. We can summarize the bag feature vectors with a set of mean map embeddings of samples in bag $b$, which is denoted here as $\bmu_b^f = \frac{1}{n_b^f}\sum_{i=1}^{n_b^f}\bphi(\x_i^{b,f}) \in \mH\in\Real^{H_f\times 1}$. The collection of all mean embeddings in the same $\mH$ is defined as 
$$\bmu_b = [\bmu_b^1,\ldots,\bmu_b^F]\in{\mathcal H},$$
and then we can use the same formula as before to define the mean map embedding, $\M=[\bmu_1|\cdots|\bmu_B]^\top\in\Real^{B\times H}$. We now need to compute the multimodal kernel matrix as follows:
\begin{eqnarray*}
\begin{array}{ll}
[\widetilde\K]_{b,b'} &= \bmu_b^\top\bmu_{b'} = 
\sum_{f=1}^F \dfrac{1}{n_b^f n_{b'}^f}\U_{n_b}^{f\top} \K_{bb'} \U_{n_{b'}^f},
\end{array}
\end{eqnarray*}
where the matrix $\K_{bb'}\in\Real^{n_b\times n_{b'}}$ is again the same size. Figure~\ref{fig:boxexamples} illustrates the two DR approaches treated in this paper.

% ---------------------------------------------------------------------------------
\subsection{Randomized distribution regression for scalability}
% ---------------------------------------------------------------------------------

The main computational bottleneck of the  distribution regression methods presented is the inversion of the $\widetilde{\bf K}$ matrix of size $B\times B$. This is a cheap operation for standard problems that involve less than a few thousand groups. However, note that each entry $(b,b')$ in matrix $\widetilde{\bf K}$ involves computing and averaging many kernel matrices and thus each entry scales in time ${\mathcal O}(n_bn_b'd)$ which reduces to ${\mathcal O}(n^2d)$ if we assume that all bags have the same number of elements $n$ (i.e. $n_b = n_b' = n$). This leads to a total cost of ${\mathcal O}( B^2n^2d + B^3)$  
where the cubic order in the number of bags $B$ is due to the matrix inversion.  
This is simply not affordable even for moderate-size problems with a few thousand points per bag. 
In this section we introduce a kernel approximation with random Fourier features~\cite{Rahimi07} that alleviates the problem.

An outstanding result in the machine learning literature makes use of a classical definition in harmonic analysis to improve approximation and scalability of kernel methods~\cite{Rahimi07}. The Bochner's theorem states that a continuous kernel $k(\x,\x')=k(\x-\x')$ on $\Real^d$ is positive definite (p.d.) if and only if $k$ is the Fourier transform of a non-negative measure. If a shift-invariant kernel $k$ is properly scaled, its Fourier transform $p({\bf w})$ is a proper probability distribution. This property is used to approximate kernel functions and matrices with linear projections on a number of $D$ random features, as follows:
\begin{align*}
k(\x, \x') & = \int_{\Real^d} p({\bf w}) e^{-\imag{\bf w}^\top({\bf x}-{\bf x}')}\mathrm{d}{\bf w} 
\approx \sum\nolimits_{i=1}^D \tfrac{1}{D} e^{-\imag{\bf w}_i^\top{\bf x}} e^{\imag{\bf w}_i^\top{\bf x}'} 
\end{align*}
where $p({\bf w})$ is set to be the inverse Fourier transform of $k$, $\imag=\sqrt{-1}$, and ${\bf w}_i \in \Real^{d}$ is randomly sampled from a data-independent distribution $p({\bf w})$~\cite{Rahimi08}. 
Note that we can define a $D$-dimensional {\em randomized} feature map ${\bf z}({\bf x}):\Real^d\to\Comp^D$, which can be explicitly constructed as ${\bf z}({\bf x}):=[\exp(\imag{\bf w}_1^\top{\bf x}),\ldots,\exp(\imag{\bf w}_D^\top{\bf x})]^\top$. Other definitions are possible: one could for instance expand the exponentials in pairs $[\cos({\bf w}_i^\top{\bf x}),\sin({\bf w}_i^\top{\bf x})]$, this increases the mapped data dimensionality to $\Real^{2D}$, while approximating exponentials by $[\cos({\bf w}_i^\top{\bf x}+b_i)]$, where $b_i\sim{\mathcal U}(0,2\pi)$, is more efficient (still mapping to $\Real^D$) but {has proved less accurate}~\cite{Sutherland15}. In our experiments we used projections onto the $[\sin,\cos]$ pairs to keep operations in the real domain. 

In matrix notation, given $n$ data points, the kernel matrix ${\bf K} \in\Real^{n\times n}$ can be approximated with the explicitly mapped data, ${\bf Z}=[{\bf z}_1\cdots{\bf z}_n]^\top\in\Real^{n\times D}$, and will be denoted as $\hat{\bf K}\approx{\bf Z}{\bf Z}^\top$. This property can be used to approximate any shift-invariant kernel. For instance, the RBF kernel, which is the one used in our experiments, can be approximated using {${\bf w}_i\sim \mathcal{N}({\bf 0}, \sigma^{-2}{\bf I})$, $1\leq i\leq D$}. 
{It is also important to notice that the approximation of $k$ with random Fourier features converges in $\ell_2$-norm error with ${\mathcal O}(D^{-1/2})$ when using an appropriate random parameter sampling distribution~\cite{Jones92}.} 

For the case of DR, in principle one should sample $B$ times (one per bag), hence obtaining $B$ sets of vectors ${\bf w}^b$ and the associated explicit maps ${\bf Z}_b$, $b=1,\ldots,B$. Such approach would be interesting if data from different bags have different dimensions. In practice, however, we will use the same random bases for all bags, thus we only sample once and obtain a single matrix ${\bf W}\in\Real^{d\times D}$. The randomized DR (RDR) readily reduces to solve the least squares regularized regression problem with the (explicit) means of projected bag data: 
$${\bf w} = ({\bf Z}^\top{\bf Z} + \lambda {\bf I}_{2D})^{-1} {\bf Z}^\top {\bf y},$$ %%w = (Mtr'*Mtr + lambda*eye(2*D))\Mtr'*ytr;
where ${\bf w}\in\Real^{2D\times 1}$ and ${\bf Z}\in\Real^{B\times 2D}$ contains the explicit mean over random Fourier feature projections per bag, that is, each row of ${\bf Z}$ contains ${\bf z}_{b}=\frac{1}{n_b}\sum_{i=1}^{n_b}{\bf z}(\x_i^b)$.  
Given a test bag dataset $\X_*\in\Real^{t\times d}$, one has to explicitly map it onto the same bases, ${\bf Z}_*=[{\bf z}(\x_1)|\cdots|{\bf z}(\x_t)]^\top\in\Real^{t\times 2D}$, compute the explicit mean map, ${\bf m}_* = \frac{1}{t}\sum_{j=1}^{t}{\bf z}(\x_j)\in\Real^{2D\times 1}$, and apply the linear prediction model on the explicit mean: 
$$\hat y_* = {\bf m}_*\w \in\Real.$$

By considering a number of samples $n \gg D $ against a moderate number of bags $B$, the associated cost by using the random features approximation now reduces to ${\mathcal O}( B^2D^2n)$. This is much more convenient than the kernel version that scales quadratically in $n$ as ${\mathcal O}( B^2n^2d)$. This situation where the number of samples is dominating the problem is the typicall scenario in real domains. 

Finally, and from the point of view of the implementation, despite having guaranteed asymptotic convergence to the RBF kernel due to the Bochner's theorem, the selection of the parameter $D$ is crucial and may act a regularizer. We cross-validated the value of $D$ to guarantee that we really captured the intrinsic dimensionality of the problem: in some cases, small $D$ values (hence not approximating the RBF kernels) led to improved performance. This effect has been previously noted in  \cite{RFF-VFF}.

%%%%%%%%%%%%%%%%%%%%%%%%%%%%%%%%%%%%%%%%%%%%%%%%%%%%%%%%%%%%%%%%%%%%%%%%%%%%%%%%%%%
\section{Experimental results} \label{sec:4}
%%%%%%%%%%%%%%%%%%%%%%%%%%%%%%%%%%%%%%%%%%%%%%%%%%%%%%%%%%%%%%%%%%%%%%%%%%%%%%%%%%%

This section shows results for several methods and in different problems of distribution regression found in remote sensing. As baseline standard approaches, we consider a least squares linear regression model (LR) and the nonlinear (kernel) counterpart, the kernel ridge regression (KR) method, both working on the means of each bag as input feature vectors. These methods will be compared with our proposals: KDR (kernel distribution regression), its randomized approximation (RDR), and in the last experiment we will add the multi-source DR (MDR) along with stacked-feature approaches. We provide empirical comparisons with standard scores like the root-mean-square-error (RMSE), coefficient of determination (R$^2$), in cross-validation and test sets to assess accuracy and robustness. We provide source code of our methods in \href{http://isp.uv.es/code/dr.html}{http://isp.uv.es/code/dr.html}.

% ---------------------------------------------------------------------------------
\subsection{Experimental setup and model development}
% ---------------------------------------------------------------------------------

The proposed methods are evaluated in three different scenarios. Firstly, SMAP Vegetation Optical Depth (VOD) data is related to crop production data from the 2015 US agricultural survey (total yield and yield per crop type). Secondly, MISR and MODIS reflectances are used to estimate Aerosol Optical Depth (AOD) (data sets in~\cite{Wang2012a}). The AOD estimation is solved using each sensor data independently, or using multi-sensor combined information through the proposed MDR formulation and a stacked-feature approach.

Evaluation of the algorithms is done as follows. We reserve a percentage of the elements of each bag for test. With the remaining elements, we perform a $k$-fold cross-validation with $k=5$ also at a bag level. This is done by randomly splitting the data into five subsets: one subset is reserved for validation and the others for training of a predictor. After this procedure, we apply the best model found to the test data. Finally, all this process is repeated ten times, and the average over all test results is computed. Both cross-validation and test errors are reported. We use as evaluation criteria the standard mean error (ME) to account for bias, the root-mean-square-error (RMSE) to assess accuracy, and the coefficient of determination or explained variance ($R^2$) to account for the goodness-of-fit. 

% ---------------------------------------------------------------------------------
\input{crop_yield_experiment.tex}

% ---------------------------------------------------------------------------------

% ---------------------------------------------------------------------------------
\input{aod_experiment.tex}
% ---------------------------------------------------------------------------------

%%%%%%%%%%%%%%%%%%%%%%%%%%%%%%%%%%%%%%%%%%%%%%%%%%%%%%%%%%%%%%%%%%%%%%%%%%%%%%%%%%%
\section{Conclusions} \label{sec:5}
%%%%%%%%%%%%%%%%%%%%%%%%%%%%%%%%%%%%%%%%%%%%%%%%%%%%%%%%%%%%%%%%%%%%%%%%%%%%%%%%%%%

This paper presented several methods based on kernels for distribution regression, and illustrated their performance in several scenarios of interest in remote sensing: 1) estimation of total crop and crop-specific yield from SMAP Vegetation Optical Depth, 2) estimation of Aerosol Optical Depth from MISR and MODIS reflectances, separately; and 3) combination of MISR and MODIS reflectances for the estimation of Aerosol Optical Depth. Our proposed methods successfully outperforms naive approaches based on input-space means and clustering, and is competitive with previously presented methods for multiple-instance learning. 

The methods can be used in many other problems in geoscience, remote sensing and environmental sciences that require predicting a scalar from a set of vectors. We foresee many potential applications of interest beyond crop yield prediction or Aerosol Optical Depth from remote sensing data: for example predicting poverty, carbon stocks, plagues and diseases, population density, wealth or urbanization at county, region or country levels from multi-temporal and multi-sensor remote sensing data, are some possible examples. Currently, we are extending the method and application to work jointly with Sentinel-1 and 2 data for crop yield prediction in Europe. Methodologically, our agenda is tied to extend the framework to deal with missing data and features. 

\bibliographystyle{IEEEtran}
\bibliography{mil_dr,target,BOOKbib,ksnr,DSPKM,newbib}

% insert where needed to balance the two columns on the last page with
% biographies
%\vfill
\begin{IEEEbiography}[{\includegraphics[width=1in,height=1.25in,keepaspectratio]{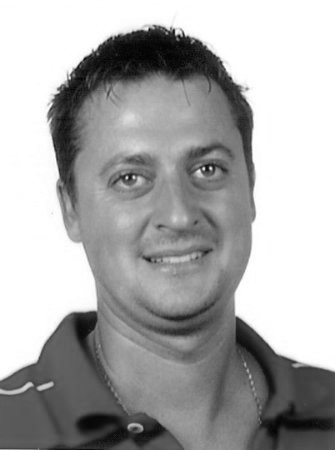}}]{Jose E. Adsuara}
received the M.Sc. degree in computer science and engineering from Universitat Polit\`{e}cnica de Catalunya (UPC), Spain. He spent a few years working in industry and education. After this, he also received the M.Sc. degree in mathematics and the Ph.D. degree in advanced physics, astrophysics, both from Universitat de Val\`{e}ncia (UV), Spain. He is currently an assistant professor in the Department of Computer Science at UV. He is also with the Image Processing Laboratory (IPL), Universitat de Val\`{e}ncia, since 2018 as a postdoctoral researcher working on machine learning for remote sensing.
\end{IEEEbiography}
%\vfill
\begin{IEEEbiography}[{\includegraphics[width=1in,height=1.25in,keepaspectratio]{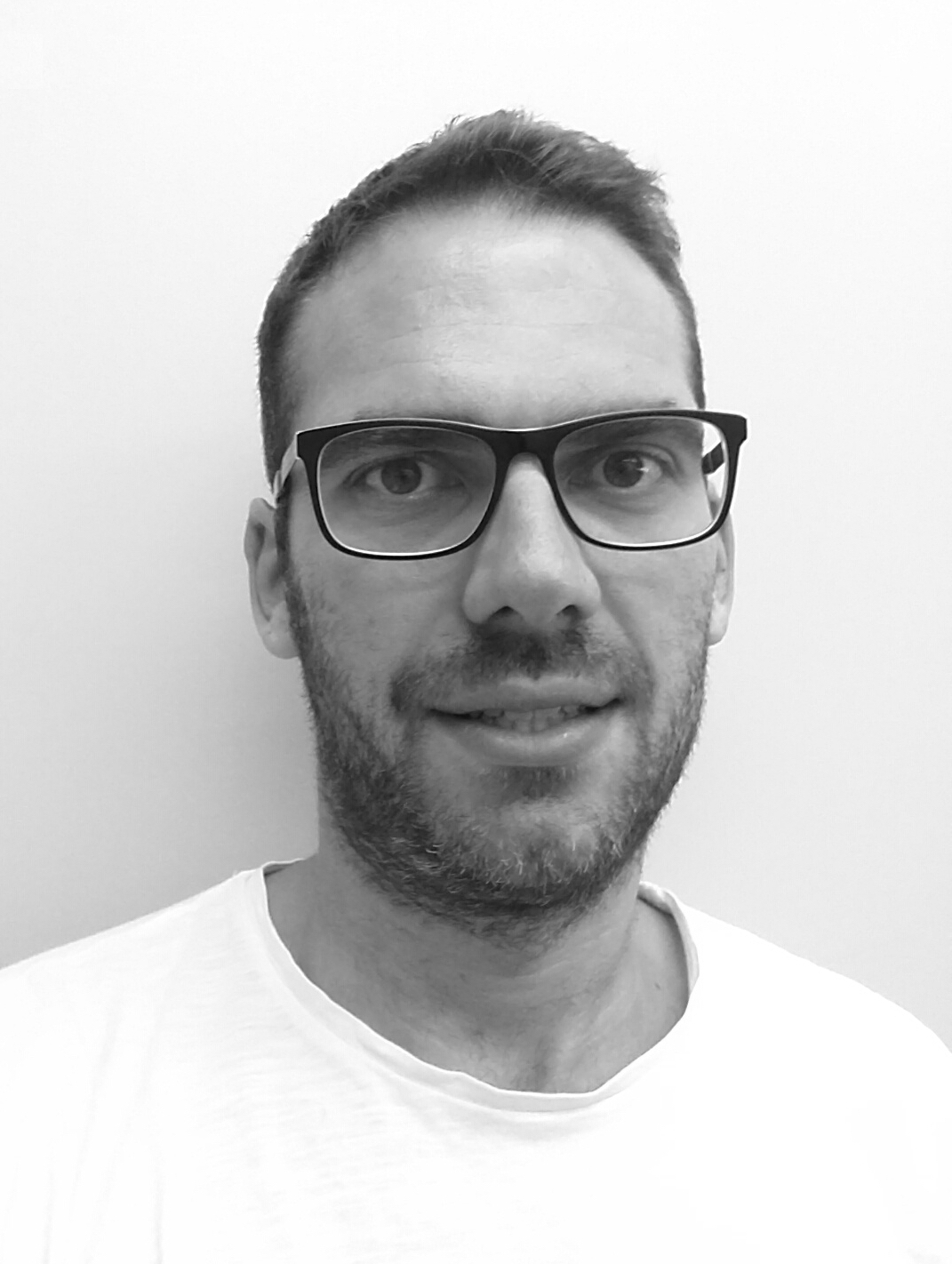}}]{Adri\'{a}n P\'{e}rez-Suay}
obtained his B.Sc. degree in Mathematics (2007), Master degree in 
Advanced Computing and Intelligent Systems (2010) and the Ph.D. degree in 
Computational Mathematics and Computer Science (2015), all from the 
Universitat de Val\`encia. He is assistant professor in the Department of 
Mathematics in the Universitat de Val\`encia. He is currently a 
Postdoctoral Researcher at the Image Processing Laboratory working 
on dependence estimation, kernel methods and causal inference for remote 
sensing data analysis.
\end{IEEEbiography}
%\vfill
\begin{IEEEbiography}[{\includegraphics[width=1in,height=1.25in,keepaspectratio]{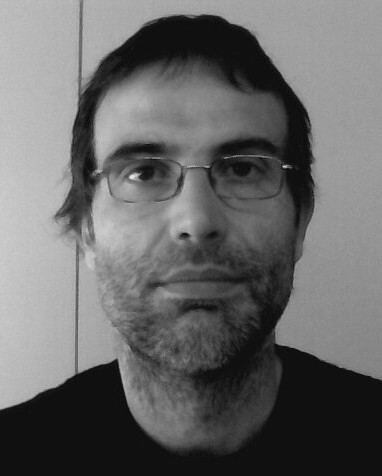}}]{Jordi Mu\~{n}oz-Mar\'{i}} 
was born in Valencia, Spain in 1970, and received a B.Sc. degree in Physics (1993), a B.Sc. degree in Electronics Engineering (1996), and a Ph.D. degree in Electronics Engineering (2003) from the Universitat de Val\`{e}ncia (UV). At present he is an Associate Professor in the Electronics Engineering Department at UV, where he teaches machine learning, big data, digital signal processing and electronics. He is a research member of the Image Processing Laboratory (https://isp.uv.es/). His research interests are tied to machine learning, statistical methods and digital signal processing applied to data analysis in general and remote sensing in particular. He is a skilled programmer in different computer languages such as C/C++, Java, Python and others. He is author and co-author of many journal papers, book chapters, and conference papers. Please visit https://www.uv.es/jordi/ for more information.
\end{IEEEbiography}
%\vfill
\begin{IEEEbiography}[{\includegraphics[width=1in,height=1.25in,keepaspectratio]{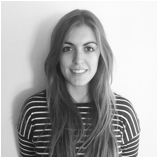}}]{Anna Mateo-Sanchis}
received the B.Sc. degree in cartography and geodesy engineering from Universitat Polit\`{e}cnica de Val\`{e}ncia (UPV), Spain. 
In 2017, she joined the Image Processing Laboratory, Universitat de Val\`{e}ncia (UV). She is currently working toward the Ph.D. degree at Universitat de Val\`{e}ncia, working on multivariate/multioutput learning methods for regression problems applied to remote sensing. Her main research interests include image processing and machine learning algorithms for Earth Observation global monitoring. 
\end{IEEEbiography}
%\vfill
\begin{IEEEbiography}[{\includegraphics[width=1in,height=1.25in,keepaspectratio]{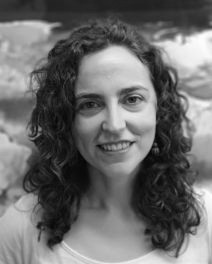}}]{Maria Piles} 
(S'05-M'11-SM'19) received the M.Sc. degree (2005) in telecommunication engineering from Universitat Polit\`ecnica de Val\`encia, Spain, and the Ph.D. degree (2010) in signal theory and communications, from the Universitat Polit\`{e}cnica de Catalunya (UPC), Spain, mastering in remote sensing.

In 2010, she was Research Fellow at University of Melbourne, Australia. From 2011 to 2015, she was Research Scientist at UPC and Affiliated Scientist at Massachusetts Institute of Technology, Cambridge. In 2016, she joined the Institute of Marine Sciences, CSIC, as a Research Scientist. Since 2017, she is with the Image Processing Laboratory, Universitat de Val\`{e}ncia, as a Ram\'on y Cajal Senior Researcher. She has wide experience in the retrieval of the water content in soils and vegetation from low-frequency microwaves, and has been actively involved within the scientific activities of the ESA's SMOS and the NASA's SMAP missions. She is a member of the CIMR Mission Advisory Group. Her research interests include microwave remote sensing, estimation of soil moisture and vegetation bio-geophysical parameters, and development of multi-sensor techniques for enhanced retrievals with focus on agriculture, forestry, wildfire prediction, extreme detection and climate studies. She is currently serving as president of the Spanish chapter of the IEEE Geoscience and Remote Sensing Society.
\end{IEEEbiography}
\vfill
\begin{IEEEbiography}[{\includegraphics[width=1in,height=1.25in,keepaspectratio]{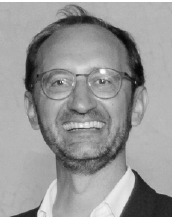}}] {Gustau Camps-Valls}
(M'04$-$SM'07$-$F'18) received the Ph.D. degree
in physics from the Universitat de Val\`encia, Valencia, Spain, in 2002.
He is currently a Full Professor of electrical engineering and a
Coordinator of the Image and Signal Processing Group, Image Processing Laboratory, with the Universitat de Val\`encia. He is involved in the development of machine learning
algorithms for geoscience and remote sensing data analysis. He has authored
200 journal papers, more than 200 conference papers, and 20 international
book chapters. He holds a Hirsch's index,h = 60 (source: Google Scholar),
entered the ISI list of Highly Cited Researchers in 2011, and Thomson
Reuters ScienceWatch identified one of his papers on Kernel-based analysis
of hyperspectral images as a Fast Moving
Front research.

In 2015, he was a recipient of the Prestigious European Research Council
(ERC) Consolidator Grant on Statistical Learning for Earth Observation Data
Analysis. He is/has been the Associate Editor of the IEEE TRANSACTIONS ON
SIGNAL PROCESSING, the IEEE GEOSCIENCE AND REMOTE SENSING LETTERS, the IEEE
SIGNAL PROCESSING LETTERS, and the
Invited Guest Editor for IEEE JOURNAL OF SELECTED TOPICS IN SIGNAL
PROCESSING in 2012 and IEEE \textit{Geoscience and Remote Sensing Magazine}
in 2015. He serves as the Editor for the books \textit{Kernel Methods
engineering}, \textit{Signal and Image Processing} (IGI, 2007),
\textit{Kernel Methods for Remote Sensing Data Analysis} (Wiley \& Sons,
2009), \textit{Remote Sensing Image Processing} (MC, 2011), and
\textit{Digital Signal Processing with Kernel Methods}
(Wiley \& Sons, 2018).
\end{IEEEbiography}
%\vfill

% You can push biographies down or up by placing
% a \vfill before or after them. The appropriate
% use of \vfill depends on what kind of text is
% on the last page and whether or not the columns
% are being equalized.

%\vfill

% Can be used to pull up biographies so that the bottom of the last one
% is flush with the other column.
%\enlargethispage{-5in}

\end{document}

%% file: crop_yield_experiment.tex
% ----------------------------------------------------------------------------------
\subsection{Estimation of crop yield}
% ----------------------------------------------------------------------------------

In this experiment, we use satellite-based retrievals of vegetation optical depth (VOD) from SMAP \cite{Konings2017} for crop-yield prediction. The vegetation optical depth (VOD) is a measure of the attenuation of soil microwave emissions when they pass through the vegetation canopy; it is sensitive to the amount of living biomass as well as to the amount of water stress experienced by the vegetation \cite{Jackson1991}. SMAP VOD has been shown to carry information about crop growth and yield in a variety of agro-ecosystems \cite{Piles2017,Chaparro18}.  

\begin{figure*}[t!]
\begin{center}
\setlength{\tabcolsep}{2pt}
\begin{tabular}{cc}
 (a) Study area & (b) Official corn yield  ($kg/m^2$) \\
\includegraphics[height=5cm]{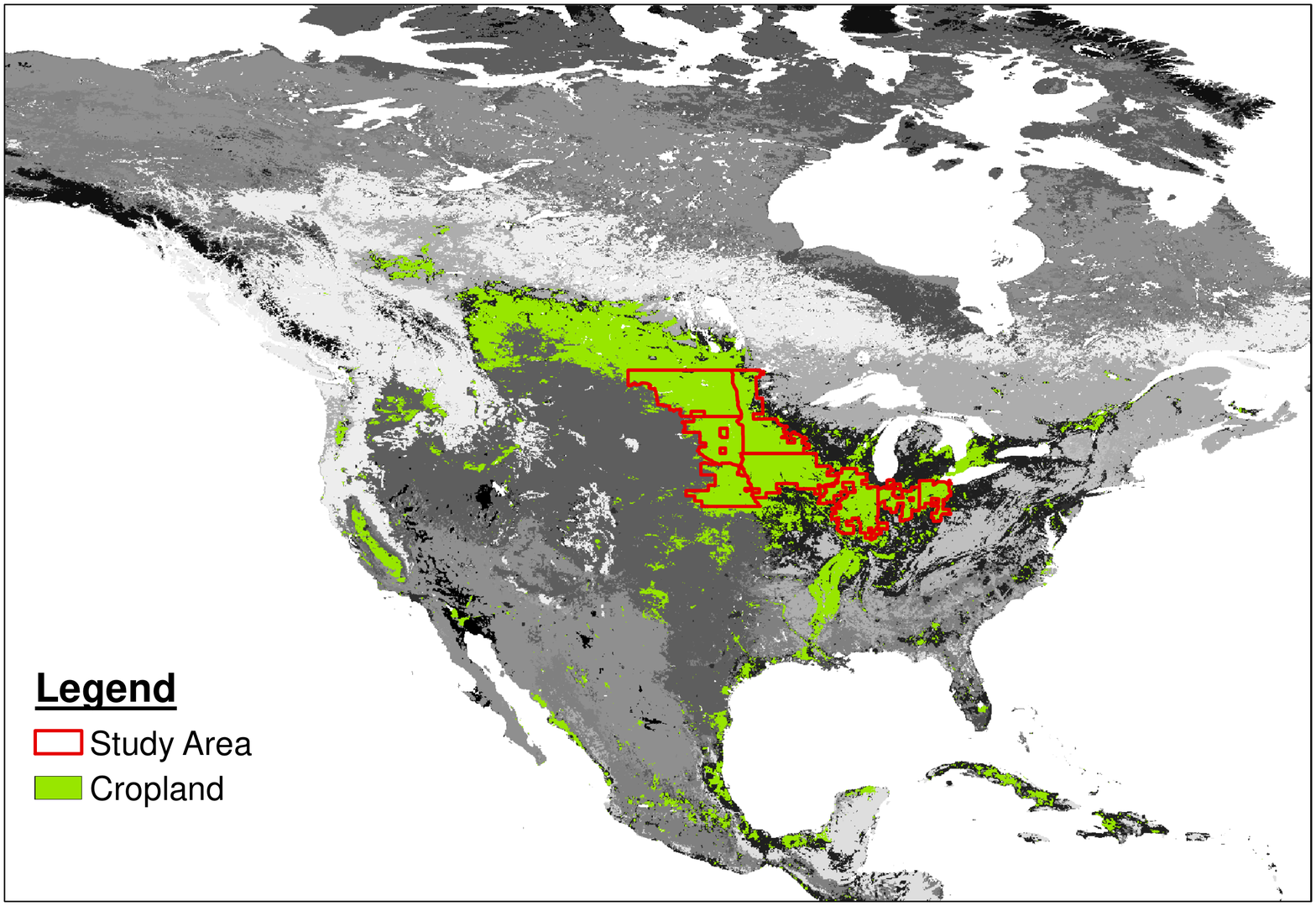} & \includegraphics[height=4cm]{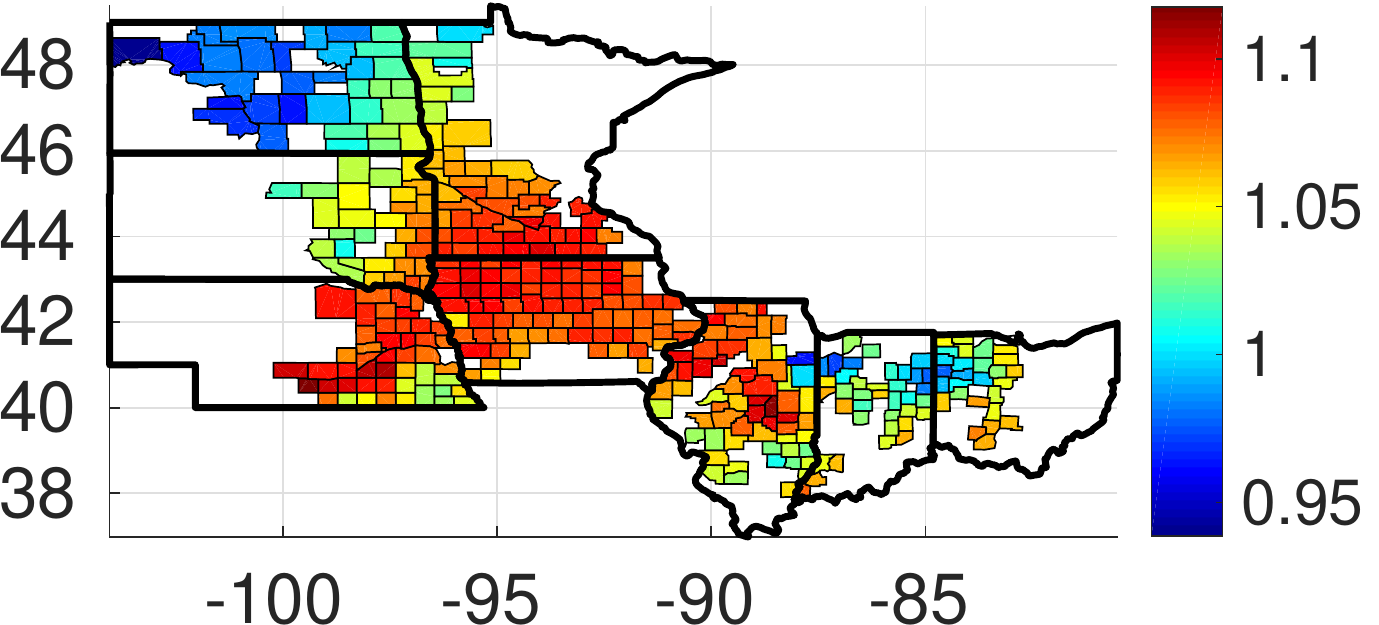}\\[.3cm]

(c) KDR relative error (\%) & (d) KDR predicted corn yield  ($kg/m^2$) \\[.2cm]
\includegraphics[height=4cm]{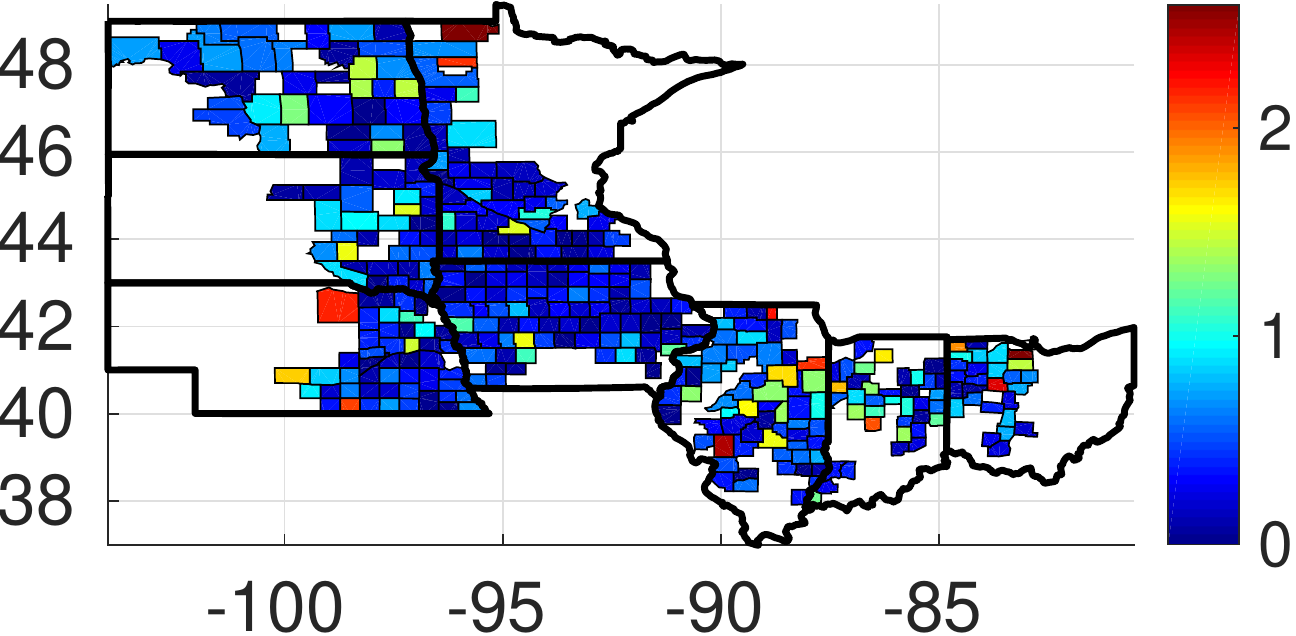} & \includegraphics[height=4cm]{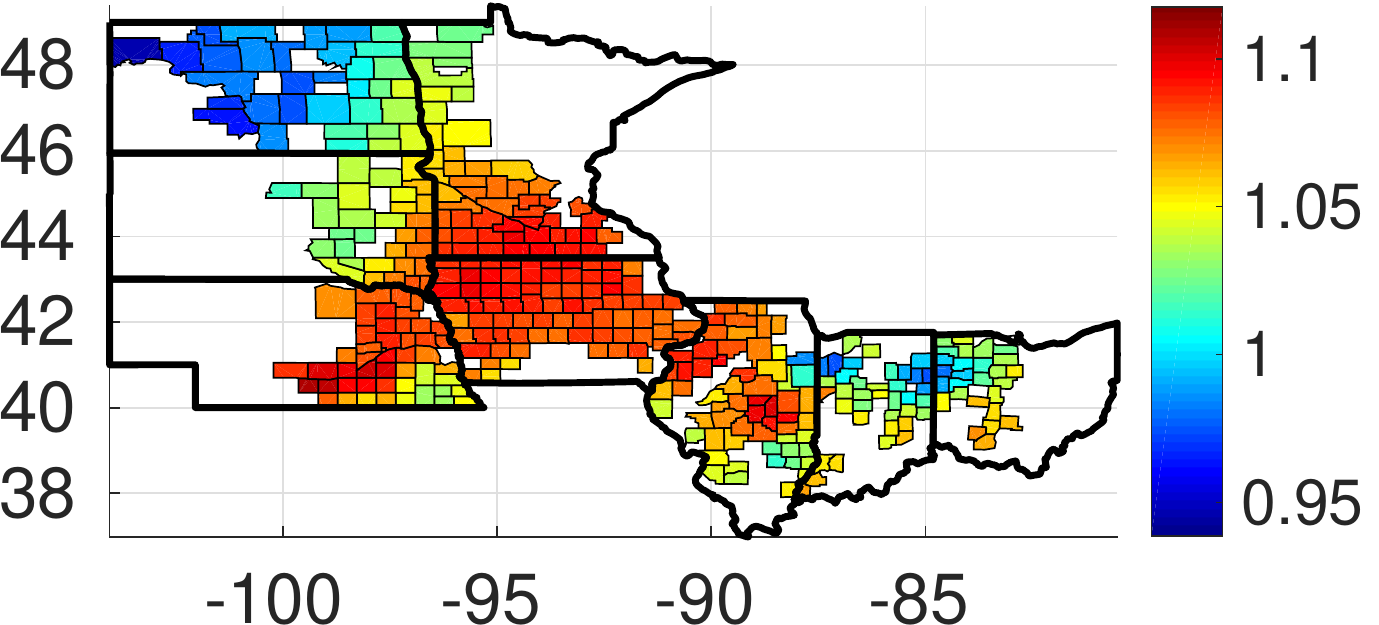}

\end{tabular}
\end{center}
\caption{(a) Study Area including the 8 states and cropland mask following the MODIS IGBP land cover classification. (b) Map of corn yield for year 2015 from USDA-NASS survey ($kg/m^2$). (c) KDR relative error prediction per county (\%). (d) KDR predicted corn yield per county ($kg/m^2$)}
\label{fig:maps}
\end{figure*}

The data set used in this study was originally presented in \cite{Chaparro18}. The study area are the extensive croplands of North Dakota, South Dakota, Nebraska, Minnesota, Iowa, Illinois, Indiana and Ohio, within the US Corn Belt (see Fig \ref{fig:maps}(a)). County-based information on area planted and yield per crop type was obtained from the US Department of Agriculture (USDA-NASS). Yields for a variety of crops are reported, namely corn, soybeans, wheat, oats, beans, barley, peas, canola, flaxseed, sorghum and lentils. A single yield datum was obtained for each county as a weighted average of the reported yield and area planted for each crop, after converting the units of each crop to $kg/m^2$ (see Table S1 in \cite{Chaparro18}). The study period is one year, starting in April 1, 2015. This corresponds to the first year of SMAP data and the 2015 crop season in the US corn belt. Only SMAP observations over agricultural pixels are considered, following the MODIS IGBP land cover classification. In \cite{Chaparro18}, the yield-VOD relationship was explored using principal components regression with VOD seasonal metrics, and the first principal component allowed explaining 66\% of variability. In this experiment, we apply the DR method to explore the yield-VOD relationship at the county scale. Each county is treated as a \emph{bag} for the DR method, containing a number of remotely sensed observations (16 SMAP pixels on average). No information on crop season is used, and the whole VOD time series are used as input. There is a total of 385 counties with yield and satellite data. A $66\%$ of these counties (bags) are used to train/validate and the remaining $33\%$ are used to test.

We show two approaches to the problem: prediction of total yield, and prediction of yield per crop type. For the latter, the three main crops in the region are predicted: corn, soybean and wheat. All the 363, 361 and 204 counties reporting corn, soybean and wheat yield, independently of their relative importance at the county level, were included in the corresponding crop-specific experiments.

Table~\ref{table:crop-yield} shows the crop-yield predictions for the baseline and proposed regression approaches. Notably, these results outperform those obtained in previous literature for corn-soy croplands (see \cite{Chaparro18} and references therein), even with the simple models (LR or KRR). This can possibly be due to the fact that we are using the whole growing-cycle information contained in the time series to develop the models, and should be a matter of future dedicated studies. Relevant to this work, consistently best results are obtained with the DR models (either RDR or KDR) in all experiments: they are able to explain 81\% of total yield, 89\% of corn yield, 86\% of soybean yield and up to 69\% of wheat yield. Everything worked as expected: the higher the capacity of the model, the better results we obtained (understanding this capacity as the treatment of the non-linearity and the grouping by bags). In particular, the advantage of non-linear %and multiple-instance-learning
approaches is more evident in the more complex prediction of independent crop yields in mixed-crops, than in the prediction of total yield, where the contributions of all crops are accounted for.

\begin{table}[t!]
\footnotesize
\caption{Results for Crop-Yield Estimation Using VOD.}
\label{table:crop-yield}
\vspace{-0.25cm}
\begin{center}
\renewcommand{\tabcolsep}{0.1cm}
\resizebox{\columnwidth}{!} {
\begin{tabular}{lc|rclrclrcl}
\hline
\hline
& {\bf Algorithms} & \multicolumn{9}{c}{\bf Total crop yield} \\
& & \multicolumn{3}{c}{ME$\times 1000$} & \multicolumn{3}{c}{RMSE $\times 100$} & \multicolumn{3}{c}{R$^2$} \\
\hline
\hline
& LR &1.19 & $\pm$ & 7.36 & 9.67 & $\pm$ & 0.74 & 0.80 & $\pm$ & 0.02  \\
& KR & 2.22 & $\pm$ & 10.77 & 9.34 & $\pm$ & 0.73 & 0.81 & $\pm$ & 0.02  \\
& RDR &2.27 & $\pm$ & 10.98 & 9.37 & $\pm$ & 0.71 & 0.81 & $\pm$ & 0.02  \\
& KDR & 2.27 & $\pm$ & 10.95 & 9.35 & $\pm$ & 0.71 & 0.81 & $\pm$ & 0.02 \\
\hline
\hline
& & \multicolumn{9}{c}{\bf Corn yield} \\
& & \multicolumn{3}{c}{ME$\times 1000$} & \multicolumn{3}{c}{RMSE $\times 100$} & \multicolumn{3}{c}{R$^2$} \\
\hline
\hline
& LR &-1.20 & $\pm$ & 5.89 & 7.54 & $\pm$ & 0.50 & 0.85 & $\pm$ & 0.02 \\
& KR &1.68 & $\pm$ & 8.52 & 6.54 & $\pm$ & 0.72 & 0.88 & $\pm$ & 0.02 \\
& RDR &1.52 & $\pm$ & 7.90 & 6.57 & $\pm$ & 0.70 & 0.88 & $\pm$ & 0.02 \\
& KDR &1.59 & $\pm$ & 7.88 & 6.47 & $\pm$ & 0.74 & 0.89 & $\pm$ & 0.02\\
\hline
\hline
& & \multicolumn{9}{c}{\bf Soybean yield} \\
& & \multicolumn{3}{c}{ME$\times 1000$} & \multicolumn{3}{c}{RMSE $\times 100$} & \multicolumn{3}{c}{R$^2$} \\
\hline
\hline
& LR &-1.99 & $\pm$ & 1.85 & 2.45 & $\pm$ & 0.13 & 0.85 & $\pm$ & 0.03 \\
& KR &-0.70 & $\pm$ & 2.92 & 2.47 & $\pm$ & 0.21 & 0.85 & $\pm$ & 0.04 \\
& RDR &-0.90 & $\pm$ & 2.58 & 2.44 & $\pm$ & 0.23 & 0.85 & $\pm$ & 0.04 \\
& KDR &-0.64 & $\pm$ & 2.43 & 2.40 & $\pm$ & 0.21 & 0.86 & $\pm$ & 0.03\\
\hline
\hline
& & \multicolumn{9}{c}{\bf Wheat yield} \\
& & \multicolumn{3}{c}{ME$\times 1000$} & \multicolumn{3}{c}{RMSE $\times 100$} & \multicolumn{3}{c}{R$^2$} \\
\hline
\hline
& LR &2.72 & $\pm$ & 6.65 & 5.46 & $\pm$ & 0.48 & 0.64 & $\pm$ & 0.08 \\
& KR &2.42 & $\pm$ & 8.47 & 5.07 & $\pm$ & 0.38 & 0.69 & $\pm$ & 0.05 \\
& RDR &3.31 & $\pm$ & 7.64 & 5.15 & $\pm$ & 0.43 & 0.68 & $\pm$ & 0.05 \\
& KDR &2.91 & $\pm$ & 7.31 & 5.10 & $\pm$ & 0.40 & 0.69 & $\pm$ & 0.05\\
\hline
\hline
\end{tabular}
}
\end{center}
\end{table}

Results of the best regression model between VOD and official corn yields at county level are illustrated in Fig.~\ref{fig:maps}. Except in some specific counties, it can be seen that the corn predictions are reasonably good, with relative errors below 3\%. We anticipate that the proposed DR approaches will be particularly useful for regional crop forecasting in areas covering different agro-climatic conditions and fragmented agricultural landscapes (e.g. Europe), where scale effects need to be properly addressed for adequate analysis and predictions \cite{LopezLozano15}.

%% file: aod_experiment.tex
% ----------------------------------------------------------------------------------
\subsection{Experiment 2: Multisensor aerosol optical depth estimation}
% ----------------------------------------------------------------------------------

This second application consists on the prediction of the Aerosol Optical Depth (AOD) from remotely-sensed data. 
{Aerosol data consist of {\em in situ} data obtained from the Aerosol Robotic Network~\cite{AERONET} (AERONET), and remote sensing data are in this case measurements from the Multi-angle Imaging SpectroRadiometer (MISR), and from the Moderate Resolution Imaging Spectroradiometer (MODIS) sensors on-board the TERRA satellite.} Aerosol estimation is one of the biggest challenges of current climate research. Aerosols both reflect and absorb incoming solar radiation, and their presence directly affects the Earth's radiation budget. Three different datasets of study are used here and originally presented in~\cite{Wang2012a}: 
\begin{itemize}
\item The AOD-MISR1 data set is a collection of $800$ bags collected at $35$ AERONET ground sites within the continental U.S. between 2001 and 2004 from the MISR sensor. Each bag consists of 100 instances, representing randomly selected pixels within 20-km radius around a given AERONET site, which leads to a total number of $76040$ instance samples. The instance attributes are 12 reflectances from the three middle MISR cameras as well as four solar and view zenith angles. The bag target value is the AOD measured by the AERONET instrument within 30 min of the satellite overpass.

\item The AOD-MISR2 data set has the same properties as AOD-MISR1. The only difference is that the 100 instances in each bag are sampled only from the cloud-free pixels. It consists on a total of 800 bags and 76040 instance samples. Since cloudy pixels are known to be noisy and lead to reduced retrieval quality, this data set is expected to provide better predictions.

\item The AOD-MODIS data set is a collection of 1364 bags collected at 45 AERONET sites within the continental U.S. between 2002 and 2004 from the MODIS sensor. Each bag consists of 100 instances. The instance attributes are seven MODIS reflectance bands and five solar and view zenith angles, and the bag label was the corresponding AERONET AOD measurement.
\end{itemize}

\begin{table}[h!] 
\footnotesize
\caption{Cross-Validation Results on the Three Remote Sensing Data Sets (See Text for Details) for AOD Prediction.}
\label{table:AOD-xval}
\vspace{-0.25cm}
\begin{center}
\renewcommand{\tabcolsep}{0.1cm}
\begin{tabular}{c|rcl|rcl|rclrcl}
\hline\hline
{\bf Algorithms} &
  \multicolumn{3}{c|}{\bf MISR1} &
  \multicolumn{3}{c|}{\bf MISR2} &
  \multicolumn{3}{c}{\bf MODIS} \\
& \multicolumn{3}{c|}{RMSE $\times 100$} &
  \multicolumn{3}{c|}{RMSE $\times 100$} &
  \multicolumn{3}{c}{RMSE $\times 100$} \\
\hline
LR  & 21.59 & $\pm$ & 0.48 & 21.01 & $\pm$ & 0.38 & 21.65 & $\pm$ & 0.37 \\
KR  &  9.07 & $\pm$ & 0.46 &  7.70 & $\pm$ & 0.76 & 11.64 & $\pm$ & 0.63 \\
RDR &  7.90 & $\pm$ & 0.50 &  7.21 & $\pm$ & 0.65 &  9.69 & $\pm$ & 0.39 \\
Best in~\cite{Wang2012a} 
    &  7.5 & $\pm$ &  0.1  &  7.3  & $\pm$ & 0.1  &  9.5  & $\pm$ & 0.1  \\
Best in~\cite{Szabo2014}
    &  7.90 & $\pm$ & 1.63 &       &       &      &       &       &      \\
\hline\hline
\end{tabular}
\end{center}
\end{table}

The way to proceed in applications of this type consists of predicting the current year AOD from the previous ones. Although the current data are not labeled per year, they correspond to four years for MISR1 and MISR2 (three years for MODIS). We reserved the equivalent of one year of data, $25\%$ of them ($33\%$ for MODIS) for testing, and the remaining $75\%$ ($66\%$ for MODIS) of data for train/validation. It is worth to note here that these sets will be different each of the $10$ times we repeat the experiment. Reported results are the average of these $10$ trials. We normalize the data. 

\begin{table*}[ht!] 
\footnotesize
\caption{Test Results on the Three Remote Sensing Data Sets (See Text for Details) for AOD Prediction.}
\label{table:AOD}
\vspace{-0.25cm}
\begin{center}
\renewcommand{\tabcolsep}{0.1cm}
\resizebox{\textwidth}{!} {
\begin{tabular}{c|rclrclrcl|rclrclrcl|rclrclrcl}
\hline\hline
{\bf Algorithms} &
  \multicolumn{9}{c|}{\bf MISR1} &
  \multicolumn{9}{c|}{\bf MISR2} &
  \multicolumn{9}{c}{\bf MODIS} \\
& \multicolumn{3}{c}{ME$\times 1000$} &
  \multicolumn{3}{c}{RMSE $\times 100$} &
  \multicolumn{3}{c|}{R$^2$} &
  \multicolumn{3}{c}{ME$\times 1000$} &
  \multicolumn{3}{c}{RMSE $\times 100$} &
  \multicolumn{3}{c|}{R$^2$} &
  \multicolumn{3}{c}{ME$\times 1000$} &
  \multicolumn{3}{c}{RMSE $\times 100$} &
  \multicolumn{3}{c}{R$^2$} \\
\hline 
LR &-6.14 & $\pm$ & 8.02 & 11.38 & $\pm$ & 0.90 & 0.64 & $\pm$ & 0.05 &-6.85 & $\pm$ & 6.28 & 9.90 & $\pm$ & 0.85 & 0.73 & $\pm$ & 0.05 &-1.64 & $\pm$ & 5.13 & 13.32 & $\pm$ & 0.92 & 0.54 & $\pm$ & 0.05 \\
KR &-2.33 & $\pm$ & 8.45 & 9.63 & $\pm$ & 0.65 & 0.74 & $\pm$ & 0.03 &-1.73 & $\pm$ & 6.50 & 8.87 & $\pm$ & 1.46 & 0.78 & $\pm$ & 0.06 &-0.75 & $\pm$ & 7.51 & 12.15 & $\pm$ & 1.16 & 0.62 & $\pm$ & 0.06 \\
RDR &-4.28 & $\pm$ & 7.40 & 8.30 & $\pm$ & 0.85 & 0.81 & $\pm$ & 0.03 &-1.71 & $\pm$ & 6.37 & 8.16 & $\pm$ & 1.26 & 0.81 & $\pm$ & 0.05 &-2.08 & $\pm$ & 5.61 & 10.09 & $\pm$ & 0.78 & 0.73 & $\pm$ & 0.06 \\
KDR &-4.34 & $\pm$ & 7.43 & 8.20 & $\pm$ & 0.83 & 0.81 & $\pm$ & 0.03 & -1.72 & $\pm$ & 6.31 & 8.18 & $\pm$ & 1.31 & 0.81 & $\pm$ & 0.05 & -2.42 & $\pm$ & 5.82 & 10.27 & $\pm$ & 1.46 & 0.73 & $\pm$ & 0.08\\
\hline\hline
\end{tabular}
}
\end{center}
\end{table*}

We show two approaches to the problem. First, we will apply the proposed DR methods using information from each sensor separately (unisensor approach), and compare results with the ones reported in~\cite{Wang2012a}, as well as to baseline regression approaches. Second, we will use the multisource DR for combining information from different sensors, specifically MISR2 and MODIS (multi-sensor approach). As this is only possible if we have land information for both sensors, the first thing we do is to build a new dataset retaining only those bags having a common land measure. Originally, we had 800 bags for MISR2 and 1364 for MODIS. After the combination, we retain only 289 bags. At this point, we want to emphasize that the combination of sensors with multiple spatial resolutions is not easy in a normal setting, and quite challenging in DR settings too. The naive approach is to adjust the two data meshes and concatenate them (feature-stacking approach), so one can use any standard regression model on this feature vector. We can do this straightforwardly because each feature is normalized. While this is practical, we are losing the information of the sensor with higher resolution. The alternative approach proposed here is to use the direct sum of kernels in our MDR model to be able to treat each input independently even if they have different resolutions and number of data points (composite approach). We will also apply our methods to the sensors separately, using only the common 289 bags, to have the single-source performance when using this particular subset of bags. We show the results for the two approaches in what follows. 

% ----------------------------------------------------------------------------------
\subsubsection{Single source approach}

Tables \ref{table:AOD-xval} and \ref{table:AOD} show the results. Table~\ref{table:AOD-xval} shows results in the training/validation set obtained by cross-validation, whereas Table~\ref{table:AOD} shows results in the unseen test set. For cross-validation we only show the RMSE, that is the criterion used for the optimization of the parameters of the models. We also show the best results obtained by Wang et al.~\cite{Wang2012a} and Szab\'o et al.~\cite{Szabo2014}. The results obtained by our methods, LR, KR and KDR, are comparable to previous proposals, although KDR does not improve previous results. It can also be seen that the composition of kernels proposed by Szab\'o in~\cite{Szabo2014} obtains similar results.

For the test results in Table~\ref{table:AOD}, which are evaluated in test data not seen when training the models, we also show the mean error (ME) and the squared Pearson correlation coefficient (R$^2$). The proposed models generalize well, and, in fact, in this most difficult case, the two proposed methods perform better than the best methods presented so far. 

% ----------------------------------------------------------------------------------
\subsubsection{Multi-source approach} 

The results are shown in Table~\ref{table:AOD-mult}. The first rows present the results of the sensors MISR2 and MODIS separately, and the last ones the results of the two possible combinations of both sensors: stacked and composite. 

On the one hand, stacking inputs makes conclusions more articulated, as results depend on the models used at a great extent. If the models used are the simplest ones, e.g LR and KR, improvements are not noticeable. On the contrary, when using DR methods, either RDR and KDR, slightly better results are obtained. However, the improvements with respect the use of the sensors separately is insignificant. On the other hand, using the composite approach, the proposed MDR always improves results with respect to all previous cases, i.e., the combination of sources is better than its individual use, and the kernel additive approach is better than the stacking one.

\begin{table}[h!] 
\footnotesize
\caption{Test Results on the Combined Dataset with 289 Common Bags for MISR2 and MODIS (See Text for Details) for AOD Prediction.}
\label{table:AOD-mult}
\vspace{-0.25cm}
\begin{center}
\renewcommand{\tabcolsep}{0.1cm}
\resizebox{\columnwidth}{!} {
\begin{tabular}{c|rclrclrcl}
\hline
\hline
{\bf Algorithms} & \multicolumn{9}{c}{\bf MISR2} \\
& \multicolumn{3}{c}{ME$\times 1000$} & \multicolumn{3}{c}{RMSE $\times 100$} & \multicolumn{3}{c}{R$^2$} \\
\hline
LR &-10.98 & $\pm$ & 5.14 & 8.13 & $\pm$ & 1.55 & 0.79 & $\pm$ & 0.05 \\
KR & 1.12 & $\pm$ & 8.03 & 6.00 & $\pm$ & 0.68 & 0.87 & $\pm$ & 0.04 \\
RDR &0.86 & $\pm$ & 8.56 & 5.86 & $\pm$ & 0.77 & 0.88 & $\pm$ & 0.04  \\
KDR &0.77 & $\pm$ & 8.55 & 5.85 & $\pm$ & 0.74 & 0.88 & $\pm$ & 0.04 \\
\hline
\hline
& \multicolumn{9}{c}{\bf MODIS} \\
& \multicolumn{3}{c}{ME$\times 1000$} & \multicolumn{3}{c}{RMSE $\times 100$} & \multicolumn{3}{c}{R$^2$} \\
\hline
LR &-7.33 & $\pm$ & 6.62 & 10.81 & $\pm$ & 1.78 & 0.61 & $\pm$ & 0.08 \\
KR &3.29 & $\pm$ & 11.83 & 10.65 & $\pm$ & 1.71 & 0.64 & $\pm$ & 0.09  \\
RDR &0.73 & $\pm$ & 9.14 & 8.74 & $\pm$ & 0.99 & 0.74 & $\pm$ & 0.06  \\
KDR & 0.88 & $\pm$ & 9.38 & 8.74 & $\pm$ & 0.96 & 0.74 & $\pm$ & 0.06  \\
\hline
\hline
& \multicolumn{9}{c}{\bf MISR2+MODIS (stack)} \\
& \multicolumn{3}{c}{ME$\times 1000$} & \multicolumn{3}{c}{RMSE $\times 100$} & \multicolumn{3}{c}{R$^2$} \\
\hline
LR &-6.76 & $\pm$ & 6.95 & 7.88 & $\pm$ & 0.80 & 0.79 & $\pm$ & 0.04 \\
KR &2.74 & $\pm$ & 6.41 & 6.06 & $\pm$ & 0.66 & 0.87 & $\pm$ & 0.04 \\
RDR &2.21 & $\pm$ & 6.35 & 5.85 & $\pm$ & 0.65 & 0.87 & $\pm$ & 0.04 \\
KDR &2.41 & $\pm$ & 6.39 & 5.83 & $\pm$ & 0.64 & 0.87 & $\pm$ & 0.04\\
\hline
\hline
& \multicolumn{9}{c}{\bf MISR2+MODIS (compo)} \\
& \multicolumn{3}{c}{ME$\times 1000$} & \multicolumn{3}{c}{RMSE $\times 100$} & \multicolumn{3}{c}{R$^2$}  \\
\hline
LR &-1.11 & $\pm$ & 9.18 & 7.17 & $\pm$ & 0.98 & 0.83 & $\pm$ & 0.03 \\
KR & 0.44 & $\pm$ & 8.45 & 6.07 & $\pm$ & 0.80 & 0.87 & $\pm$ & 0.04 \\
RDR & 0.72 & $\pm$ & 7.44 & 5.72 & $\pm$ & 0.61 & 0.89 & $\pm$ & 0.03 \\
KDR & 0.70 & $\pm$ & 7.28 & 5.69 & $\pm$ & 0.62 & 0.89 & $\pm$ & 0.03\\
\hline
\hline
\end{tabular}
}
\end{center}
\end{table}

%% file: main.bbl
\begin{thebibliography}{10}
\providecommand{\url}[1]{#1}
\csname url@rmstyle\endcsname
\providecommand{\newblock}{\relax}
\providecommand{\bibinfo}[2]{#2}
\providecommand\BIBentrySTDinterwordspacing{\spaceskip=0pt\relax}
\providecommand\BIBentryALTinterwordstretchfactor{4}
\providecommand\BIBentryALTinterwordspacing{\spaceskip=\fontdimen2\font plus
\BIBentryALTinterwordstretchfactor\fontdimen3\font minus
  \fontdimen4\font\relax}
\providecommand\BIBforeignlanguage[2]{{%
\expandafter\ifx\csname l@#1\endcsname\relax
\typeout{** WARNING: IEEEtran.bst: No hyphenation pattern has been}%
\typeout{** loaded for the language `#1'. Using the pattern for}%
\typeout{** the default language instead.}%
\else
\language=\csname l@#1\endcsname
\fi
#2}}

\bibitem{Liang08}
S.~Liang, \emph{Advances in Land Remote Sensing: System, Modeling, Inversion
  and Applications}.\hskip 1em plus 0.5em minus 0.4em\relax Germany: Springer
  Verlag, 2008.

\bibitem{rodgers00}
C.~D. Rodgers, \emph{Inverse Methods for Atmospheric Sounding: Theory and
  Practice}.\hskip 1em plus 0.5em minus 0.4em\relax World Scientific Publishing
  Co. Ltd., 2000.

\bibitem{CampsValls11mc}
G.~Camps-Valls, D.~Tuia, L.~G\'omez-Chova, and J.~Malo, Eds., \emph{Remote
  Sensing Image Processing}.\hskip 1em plus 0.5em minus 0.4em\relax Morgan \&
  Claypool, Sept 2011.

\bibitem{verrelst12b}
J.~Verrelst, L.~Alonso, G.~Camps-Valls, J.~Delegido, and J.~Moreno, ``Retrieval
  of vegetation biophysical parameters using {G}aussian process techniques,''
  \emph{IEEE Trans. Geosc. Rem. Sens.}, vol.~50, no. 5 PART 2, pp. 1832--1843,
  2012.

\bibitem{Ratle20102271}
F.~Ratle, G.~Camps-Valls, and J.~Weston, ``Semisupervised neural networks for
  efficient hyperspectral image classification,'' \emph{IEEE Transactions on
  Geoscience and Remote Sensing}, vol.~48, no.~5, pp. 2271--2282, 2010, cited
  By 59.

\bibitem{Tramontana2015360}
G.~Tramontana, K.~Ichii, G.~Camps-Valls, E.~Tomelleri, and D.~Papale,
  ``Uncertainty analysis of gross primary production upscaling using random
  forests, remote sensing and eddy covariance data,'' \emph{Remote Sensing of
  Environment}, vol. 168, pp. 360--373, 2015, cited By 1.

\bibitem{Rojo17dspkm}
J.~Rojo-\'{A}lvarez, M.~Mart\'{i}nez-Ram\'{o}n, J.~Mu{\~n}oz-Mar\'{i}, and
  G.~Camps-Valls, \emph{Digital Signal Processing with Kernel Methods}.\hskip
  1em plus 0.5em minus 0.4em\relax UK: Wiley \& Sons, Apr 2017.

\bibitem{CampsValls16grsm}
G.~Camps-Valls, J.~Verrelst, J.~Muñoz-Marí, V.~Laparra, F.~Mateo-Jimenez, and
  J.~Gomez-Dans, ``A survey on gaussian processes for earth observation data
  analysis: A comprehensive investigation,'' \emph{IEEE Geoscience and Remote
  Sensing Magazine}, no.~6, June 2016.

\bibitem{warpedGP}
A.~Mateo-Sanchis, J.~Mu{\~n}oz-Mar{\'i}, A.~P{\'e}rez-Suay, and G.~Camps-Valls,
  ``Warped gaussian processes in remote sensing parameter estimation and causal
  inference,'' \emph{IEEE Geoscience and Remote Sensing Letters}, pp. 1--5,
  2018.

\bibitem{LopezLozano15}
R.~L{\'o}pez-Lozano, G.~Duveiller, L.~Seguini, M.~Meroni, S.~García-Condado,
  J.~Hooker, O.~Leo, and B.~Baruth, ``Towards regional grain yield forecasting
  with 1km-resolution eo biophysical products: Strengths and limitations at
  pan-european level,'' \emph{Agricultural and Forest Meteorology}, vol. 206,
  pp. 12 -- 32, 2015.

\bibitem{Galidaki2017}
G.~Galidaki, D.~Zianis, I.~Gitas, K.~Radoglou, V.~Karathanassi,
  M.~Tsakiri–Strati, I.~Woodhouse, and G.~Mallinis, ``Vegetation biomass
  estimation with remote sensing: focus on forest and other wooded land over
  the mediterranean ecosystem,'' \emph{International Journal of Remote
  Sensing}, vol.~38, no.~7, pp. 1940--1966, 2017.

\bibitem{Kucera2013}
P.~A. Kucera, E.~E. Ebert, F.~J. Turk, V.~Levizzani, D.~Kirschbaum, F.~J.
  Tapiador, A.~Loew, and M.~Borsche, ``Precipitation from space: Advancing
  earth system science,'' \emph{Bulletin of the American Meteorological
  Society}, vol.~94, no.~3, pp. 365--375, 2013.

\bibitem{Crow2012}
W.~T. Crow, A.~A. Berg, M.~H. Cosh, A.~Loew, B.~P. Mohanty, R.~Panciera,
  P.~de~Rosnay, D.~Ryu, and J.~P. Walker, ``Upscaling sparse ground‐based
  soil moisture observations for the validation of coarse‐resolution
  satellite soil moisture products,'' \emph{Reviews of Geophysics}, vol.~50,
  no.~2, 2012.

\bibitem{Tang2014}
W.~Tang, S.~H. Yueh, A.~G. Fore, and A.~Hayashi, ``Validation of aquarius sea
  surface salinity with in situ measurements from argo floats and moored
  buoys,'' \emph{Journal of Geophysical Research: Oceans}, vol. 119, no.~9, pp.
  6171--6189, 2014.

\bibitem{Wang2008}
Z.~Wang, ``{Aerosol Optical Depth Prediction from Satellite Observations by
  Multiple Instance Regression},'' \emph{SIAM International Conference on Data
  Mining, SIAM}, pp. 165--176, 2008.

\bibitem{Charkovska2018}
N.~Charkovska, J.~Horabik-Pyzel, R.~Bun, O.~Danylo, Z.~Nahorski, M.~Jonas, and
  X.~Xiangyang, ``High-resolution spatial distribution and associated
  uncertainties of greenhouse gas emissions from the agricultural sector,''
  \emph{Mitigation and Adaptation Strategies for Global Change}, Jan 2018.

\bibitem{Moser2015}
G.~Moser, M.~D. Martino, and S.~B. Serpico, ``Estimation of air surface
  temperature from remote sensing images and pixelwise modeling of the
  estimation uncertainty through support vector machines,'' \emph{IEEE Journal
  of Selected Topics in Applied Earth Observations and Remote Sensing}, vol.~8,
  no.~1, pp. 332--349, Jan 2015.

\bibitem{Goovaerts2010CombiningAA}
P.~Goovaerts, ``Combining areal and point data in geostatistical interpolation:
  Applications to soil science and medical geography.'' \emph{Mathematical
  geosciences}, vol. 42 5, pp. 535--554, 2010.

\bibitem{Bolton11}
J.~Bolton and P.~Gader, ``Application of multiple-instance learning for
  hyperspectral image analysis,'' \emph{IEEE Geoscience and Remote Sensing
  Letters}, vol.~8, no.~5, pp. 889--893, Sept 2011.

\bibitem{Manandhar15}
A.~Manandhar, P.~A. Torrione, L.~M. Collins, and K.~D. Morton,
  ``Multiple-instance hidden markov model for gpr-based landmine detection,''
  \emph{IEEE Transactions on Geoscience and Remote Sensing}, vol.~53, no.~4,
  pp. 1737--1745, April 2015.

\bibitem{Jiao15}
C.~Jiao and A.~Zare, ``Functions of multiple instances for learning target
  signatures,'' \emph{IEEE Transactions on Geoscience and Remote Sensing},
  vol.~53, no.~8, pp. 4670--4686, Aug 2015.

\bibitem{Yuksel15}
S.~E. Yuksel, J.~Bolton, and P.~Gader, ``Multiple-instance hidden markov models
  with applications to landmine detection,'' \emph{IEEE Transactions on
  Geoscience and Remote Sensing}, vol.~53, no.~12, pp. 6766--6775, Dec 2015.

\bibitem{Liu18}
X.~Liu, L.~Jiao, J.~Zhao, J.~Zhao, D.~Zhang, F.~Liu, S.~Yang, and X.~Tang,
  ``Deep multiple instance learning-based spatial-spectral classification for
  pan and ms imagery,'' \emph{IEEE Transactions on Geoscience and Remote
  Sensing}, vol.~56, no.~1, pp. 461--473, Jan 2018.

\bibitem{Wagstaff2007}
K.~L. Wagstaff and T.~Lane, ``{Salience Assignment for Multiple-Instance
  Regression},'' \emph{ICML '2007 Workshop on Constrained Optimization and
  Structured Output Spaces}, 2007.

\bibitem{Wang2012a}
Z.~Wang, L.~Lan, and S.~Vucetic, ``{Mixture model for multiple instance
  regression and applications in remote sensing},'' \emph{IEEE Transactions on
  Geoscience and Remote Sensing}, vol.~50, no.~6, pp. 2226--2237, 2012.

\bibitem{pmlr-v31-poczos13a}
B.~Poczos, A.~Singh, A.~Rinaldo, and L.~Wasserman, ``Distribution-free
  distribution regression,'' in \emph{Proceedings of the Sixteenth
  International Conference on Artificial Intelligence and Statistics}, ser.
  Proceedings of Machine Learning Research, C.~M. Carvalho and P.~Ravikumar,
  Eds., vol.~31.\hskip 1em plus 0.5em minus 0.4em\relax Scottsdale, Arizona,
  USA: PMLR, 29 Apr--01 May 2013, pp. 507--515.

\bibitem{pmlr-v28-oliva13}
J.~Oliva, B.~Poczos, and J.~Schneider, ``Distribution to distribution
  regression,'' in \emph{Proceedings of the 30th International Conference on
  Machine Learning}, ser. Proceedings of Machine Learning Research, S.~Dasgupta
  and D.~McAllester, Eds., vol.~28, no.~3.\hskip 1em plus 0.5em minus
  0.4em\relax Atlanta, Georgia, USA: PMLR, 17--19 Jun 2013, pp. 1049--1057.

\bibitem{pmlr-v33-oliva14a}
J.~Oliva, W.~Neiswanger, B.~Poczos, J.~Schneider, and E.~Xing, ``{Fast
  Distribution To Real Regression},'' in \emph{Proceedings of the Seventeenth
  International Conference on Artificial Intelligence and Statistics}, ser.
  Proceedings of Machine Learning Research, S.~Kaski and J.~Corander, Eds.,
  vol.~33.\hskip 1em plus 0.5em minus 0.4em\relax Reykjavik, Iceland: PMLR,
  22--25 Apr 2014, pp. 706--714.

\bibitem{Szabo2014}
Z.~Szab{{\'o}}, B.~K. Sriperumbudur, B.~P{{\'o}}czos, and A.~Gretton,
  ``Learning theory for distribution regression,'' \emph{Journal of Machine
  Learning Research}, vol.~17, no. 152, pp. 1--40, 2016.

\bibitem{Law2017}
H.~C.~L. Law, D.~J. Sutherland, D.~Sejdinovic, and S.~Flaxman, ``{Bayesian
  Approaches to Distribution Regression},'' \emph{ArXiv e-prints}, 2017.

\bibitem{Scholkopf02}
B.~Sch\"olkopf and A.~Smola, \emph{Learning with Kernels -- Support Vector
  Machines, Regularization, Optimization and Beyond}.\hskip 1em plus 0.5em
  minus 0.4em\relax MIT Press Series, 2002.

\bibitem{ShaweTaylor04}
J.~Shawe-Taylor and N.~Cristianini, \emph{Kernel Methods for Pattern
  Analysis}.\hskip 1em plus 0.5em minus 0.4em\relax New York, NY, USA:
  Cambridge University Press, 2004.

\bibitem{CampsValls09}
G.~Camps-Valls and L.~Bruzzone, Eds., \emph{Kernel methods for Remote Sensing
  Data Analysis}.\hskip 1em plus 0.5em minus 0.4em\relax UK: Wiley \& Sons,
  Dec. 2009.

\bibitem{Muandet2016}
K.~Muandet, K.~Fukumizu, B.~Sriperumbudur, and B.~Sch{\"{o}}lkopf,
  \emph{{Kernel Mean Embedding of Distributions: A Review and Beyond}}, 2016.

\bibitem{Aronszajn50}
N.~Aronszajn, ``Theory of reproducing kernels,'' \emph{Trans. Amer. Math.
  Soc.}, vol.~68, no.~3, pp. 337--404, May 1950.

\bibitem{RieNag55}
F.~Riesz and B.~S. Nagy, \emph{Functional Analysis}.\hskip 1em plus 0.5em minus
  0.4em\relax Frederick Ungar Publishing, 1955.

\bibitem{Kimeldorf1971}
G.~Kimeldorf and G.~Wahba, ``Some results on tchebycheffian spline functions,''
  \emph{Journal of Mathematical Analysis and Applications}, vol.~33, no.~1, pp.
  82--95, 1971.

\bibitem{Harchaoui13}
Z.~Harchaoui, F.~Bach, O.~Cappe, and E.~Moulines, ``Kernel-based methods for
  hypothesis testing: A unified view,'' \emph{IEEE Signal Processing Magazine},
  vol.~30, no.~4, pp. 87--97, July 2013.

\bibitem{CampsValls10hsic}
G.~Camps-Valls, J.~Mooij, and B.~Sch{\"o}lkopf, ``Remote sensing feature
  selection by kernel dependence estimation,'' \emph{IEEE Geoscience and Remote
  Sensing Letters}, vol.~7, no.~3, pp. 587--591, July 2010.

\bibitem{Matasci15}
G.~Matasci, M.~Volpi, M.~Kanevski, L.~Bruzzone, and D.~Tuia, ``Semisupervised
  transfer component analysis for domain adaptation in remote sensing image
  classification,'' \emph{IEEE Transactions on Geoscience and Remote Sensing},
  vol.~53, no.~7, pp. 3550--3564, July 2015.

\bibitem{Persello16}
C.~Persello and L.~Bruzzone, ``Kernel-based domain-invariant feature selection
  in hyperspectral images for transfer learning,'' \emph{IEEE Transactions on
  Geoscience and Remote Sensing}, vol.~54, no.~5, pp. 2615--2626, May 2016.

\bibitem{Rahimi07}
A.~Rahimi and B.~Recht, ``Random features for large-scale kernel machines,'' in
  \emph{Proceedings of the 20th International Conference on Neural Information
  Processing Systems}, ser. NIPS'07.\hskip 1em plus 0.5em minus 0.4em\relax
  USA: Curran Associates Inc., 2007, pp. 1177--1184.

\bibitem{Rahimi08}
A.~{Rahimi} and B.~Recht, ``Weighted sums of random kitchen sinks: Replacing
  minimization with randomization in learning,'' in \emph{Advances in Neural
  Information Processing Systems 21}, D.~Koller, D.~Schuurmans, Y.~Bengio, and
  L.~Bottou, Eds., 2009, pp. 1313--1320.

\bibitem{Sutherland15}
J.~Sutherland and J.~Schneider, ``On the error of random fourier features,'' in
  \emph{UAI}, 2015, pp. 862--871.

\bibitem{Jones92}
L.~K. Jones, ``Annals of statistics,'' \emph{A simple lemma on greedy
  approximation in {H}ilbert space and convergence rates for projection pursuit
  regression and neural network training}, vol.~20, pp. 608--613, 1992.

\bibitem{RFF-VFF}
P.~Morales-{\'A}lvarez, A.~P{\'e}rez-Suay, R.~Molina, and G.~Camps-Valls,
  ``Remote sensing image classification with large-scale gaussian processes,''
  \emph{IEEE Transactions on Geoscience and Remote Sensing}, vol.~56, no.~2,
  pp. 1103--1114, Feb 2018.

\bibitem{Konings2017}
A.~G. Konings, M.~Piles, N.~Das, and D.~Entekhabi, ``L-band vegetation optical
  depth and effective scattering albedo estimation from smap,'' \emph{Remote
  Sensing of Environment}, vol. 198, pp. 460 -- 470, 2017.

\bibitem{Jackson1991}
T.~Jackson and T.~Schmugge, ``Vegetation effects on the microwave emission of
  soils,'' \emph{Remote Sensing of Environment}, vol.~36, no.~3, pp. 203 --
  212, 1991.

\bibitem{Piles2017}
M.~Piles, G.~Camps-Valls, D.~Chaparro, D.~Entekhabi, A.~G. Konings, and
  T.~Jagdhuber, ``Remote sensing of vegetation dynamics in agro-ecosystems
  using smap vegetation optical depth and optical vegetation indices,'' in
  \emph{IGARSS17}, July 2017, pp. 4346--4349.

\bibitem{Chaparro18}
D.~Chaparro, M.~Piles, M.~Vall-llossera, A.~Camps, A.~G. Konings, and
  D.~Entekhabi, ``L-band vegetation optical depth seasonal metrics for crop
  yield assessment,'' \emph{Remote Sensing of Environment}, vol. 212, pp.
  249--259, June 2018.

\bibitem{AERONET}
B.~Holben, T.~Eck, I.~Slutsker, D.~Tanré, J.~Buis, A.~Setzer, E.~Vermote,
  J.~Reagan, Y.~Kaufman, T.~Nakajima, F.~Lavenu, I.~Jankowiak, and A.~Smirnov,
  ``Aeronet—a federated instrument network and data archive for aerosol
  characterization,'' \emph{Remote Sensing of Environment}, vol.~66, no.~1, pp.
  1 -- 16, 1998.

\end{thebibliography}
